\pdfoutput=1

\documentclass[11pt]{article}
\usepackage{acl}

\usepackage{times}
\usepackage{latexsym}
\usepackage{multirow}

\usepackage[T1]{fontenc}
\PassOptionsToPackage{table,xcdraw}{xcolor}
\usepackage[table,xcdraw]{xcolor}
\usepackage{colortbl}

\usepackage[utf8]{inputenc}
\usepackage{microtype}
\usepackage{hyperref}
\usepackage{graphicx}
\usepackage{amsmath}
\usepackage{babel}
\usepackage{booktabs}
\usepackage{array}
\usepackage{caption}
\usepackage{pifont}
\usepackage{amssymb}
\usepackage{arydshln} 
\usepackage{amsmath, amssymb, amsthm, enumerate, graphicx}
\usepackage{url}
\usepackage{enumitem}
\usepackage{newtxtext,newtxmath}

\newcommand{\dataset}{\texttt{ANGST}}
\newcommand{\Silverdataset}{\texttt{ANGST-SILVER}}
\newcommand\blfootnote[1]{%
  \begingroup
  \renewcommand\thefootnote{}\footnote{#1}%
  \addtocounter{footnote}{-1}%
  \endgroup
}

\definecolor{yellow}{HTML}{FFB300}
\definecolor{blue}{HTML}{2196F3}
\definecolor{green}{HTML}{689F38}
\definecolor{red}{HTML}{EF5350}
\definecolor{ForestGreen}{RGB}{34,139,34}
\definecolor{navyblue}{rgb}{0.0, 0.0, 0.5} 

\title{\textit{Still Not Quite There!} Evaluating Large Language Models for Comorbid Mental Health Diagnosis}

\author{
 \textbf{Amey Hengle \textsuperscript{$\varheartsuit\ast$}}\quad
 \textbf{Atharva Kulkarni\textsuperscript{$\spadesuit\ast$}}\quad
 \textbf{Shantanu Patankar \textsuperscript{$\clubsuit$}}
 \\
 \textbf{Madhumitha Chandrasekaran\textsuperscript{$\heartsuit$}}\quad
 \textbf{Sneha D'Silva\textsuperscript{$\varspadesuit$}}\quad
  \textbf{Jemima Jacob\textsuperscript{$\varclubsuit$}}\quad
 \textbf{Rashmi Gupta\textsuperscript{$\diamondsuit$}}
\\
 \textsuperscript{$\varheartsuit$}Skit.ai\quad
 \textsuperscript{$\spadesuit$}University of Southern California \quad
 \textsuperscript{$\clubsuit$}Georgia Institute of Technology 
 \\
 \textsuperscript{$\heartsuit$}Sion Hospital\quad
 \textsuperscript{$\varspadesuit$}Clinical Psychotherapist\quad
 \textsuperscript{$\varclubsuit$}Sophia College for Women
 \\
\textsuperscript{$\diamondsuit$}Indian Institute of Technology Bombay
\\
\small{\textbf{Correspondence:} \href{mailto:ameyhengle@gmail.com}{ameyhengle@gmail.com}
}
}

\begin{document}
\maketitle

\begin{abstract}

\textcolor{red}{\textbf{Warning:} This paper includes examples displaying symptoms of mental health disorders for contextual understanding.}

In this study, we introduce \dataset, a novel, first of its kind benchmark for depression-anxiety comorbidity classification from social media posts. Unlike contemporary datasets that often oversimplify the intricate interplay between different mental health disorders by treating them as isolated conditions, \dataset\ enables multi-label classification, allowing each post to be simultaneously identified as indicating depression and/or anxiety. Comprising $2876$ meticulously annotated posts by expert psychologists and an additional $7667$ silver-labeled posts, \dataset\ posits a more representative sample of online mental health discourse. Moreover, we benchmark \dataset\ using various state-of-the-art language models, ranging from Mental-BERT to GPT-4. Our results provide significant insights into the capabilities and limitations of these models in complex diagnostic scenarios. While GPT-4 generally outperforms other models, none achieve an F1 score exceeding $72\%$ in multi-class comorbid classification, underscoring the ongoing challenges in applying language models to mental health diagnostics.


\end{abstract}
\section{Introduction}
\blfootnote{$^\ast$Equal contribution}


With the advancement of web technologies, social media platforms such as Reddit \cite{gkotsis-etal-2016-language, gaur2018let}, Twitter \cite{de-choudhary-etal-2013, coppersmith-etal-2014-quantifying}, and ReachOut \cite{shandley2010evaluation, kahl2020evaluation}, become popular hubs for mental health support and information exchange. They offer an anonymous, safe space that fosters a sense of community and empowerment that begets in-depth mental health discussions \cite{de2014mental, berry2017whywetweetmh}. Consequently, there is a growing body of research on "\textit{digital psychiatry}" \cite{tsugawa-etal-2015} that analyzes the mental health discourse and language usage on these platforms to enhance the discovery, understanding, and detection of mental health concerns. However, despite significant efforts, there exist numerous concerns regarding the curation of these resources.

\paragraph{Drawbacks of Existing Data Resources.}

The prevalent modus operandi for sourcing mental health-related datasets involves crawling social media posts that are either intrinsic to specific online communities pertaining to mental health \cite{turcan-mckeown-2019-dreaddit} or those that bear certain attributive words \cite{mowery-etal-2016-towards}, hashtags \cite{berry2017whywetweetmh}, or self-reported diagnoses \cite{coppersmith-etal-2015-clpsych, yates-etal-2017-depression}. The binary indicator of the presence or absence of these signals is used to determine the positive class, while the negative class\footnote{In this study, we refer to the `control group' more from a machine learning perspective -- as the `negative class,'. That is, `control group' represents posts that do not contain the target labels (depression or anxiety).} constitutes randomly crawled posts. This strategy engenders highly biased data and skewed annotations incongruous with the real-world distribution. Furthermore, it exacerbates the semantic divergence between the positive and negative classes, simplifying the task and diminishing its utility in real-world settings. We term this issue as the \textbf{`\textit{Data Source Bottleneck}'}.

Moreover, these `\textit{proxy diagnostic signals}' of affiliation behavior (hashtags/community membership) or self-reports of diagnosis provide an easy and inexpensively means of collecting silver-labeled data, obviating the need for hiring professional annotators. However, as elucidated by \citet{ernala2019methodological}, they lack clinical grounding, theoretical contextualization, and psychometric validity. They presents a rather myopic approach to labeling data as it does not consider the text's semantics and focuses solely on attribution. As unearthed by \citet{resnik-etal-2013-using} and \citet{kulkarni-etal-2021-cluster}, people post a variety of content in mental health-related communities, ranging from trauma episodes and emotion regulation to general advice and therapy experiences. Thus, conglomerating them under a single label leads to noisy annotations, spurious correlations, and unwanted biases. We term this quandary the \textbf{`\textit{Annotation Bottleneck}'}.

Lastly, most existing works treat mental health disorder identification as binary classification tasks -- differentiating posts exhibiting symptoms of a particular mental illness from those of the control users. While some endeavors have been made towards multi-class classification of multiple mental health disorders \cite{cohan-etal-2018-smhd, sbbd, garg-etal-2022-cams, ji2022suicidala}, they fail to capture the main narrative -- Mental health conditions are not mutually exclusive. In reality, conditions like depression and anxiety often manifest concurrently, and a social media post may reflect both conditions simultaneously, necessitating multiple labels. For example, \citet{de-souza-etal-2022} identify persistent traits of anxiety in the Reddit posts of individuals experiencing depression. This phenomenon, known as comorbidity, is well-documented in psychology research \cite{hirschfeld2001comorbidity, lamers2011comorbidity}.
By overlooking this crucial aspect, current frameworks oversimplify the complex interplay of mental health disorders, thereby limiting our ability to develop comprehensive solutions for this multifaceted challenge. We term this drawback the \textbf{`\textit{Task Bottleneck}'}.

\paragraph{Proposed Dataset.}  To this end, we introduce \textbf{\dataset} (\textbf{AN}xiety-Depression Comorbidity Dia\textbf{G}nosis in Reddit Po\textbf{ST}), a novel corpus that addresses the \textit{data source}, \textit{annotation}, and \textit{task bottleneck} issue discussed above. Specifically, we frame comorbidity identification as a multi-label classification task, wherein a post can be labeled as indicative of depression, anxiety, both, or neither. Likewise, \dataset\ can be used for traditional binary classification of control vs mental health group. \dataset\ comprises $2876$ Reddit posts meticulously annotated by $3$ expert psychologists. Our scrupulous data acquisition and filtering pipeline ensures that \dataset\ represents data proximate to the real-world distribution of online mental health discourse. The depression, anxiety, and control class samples exhibit reduced semantic divergence, rendering them more challenging to distinguish than in contemporary datasets and, thus, better reflecting the nuances of real-world data. Furthermore, we also compile \Silverdataset, a silver-labeled corpus of $7667$ Reddit posts for comorbidity classification. This supplementary dataset is sourced from the same distribution as \dataset, but its annotations are derived through a carefully tuned prompting technique using GPT-3.5-turbo \cite{ouyang2022-instuctGPT}. It provides an ancillary resource for exploring semi-supervised and few-shot learning paradigms in the domain of mental health comorbidity analysis.

\paragraph{Benchmarking Methods.} We benchmark \dataset, on a host of discriminative pretrained models (PLMs) like Mental-BERT, Mental-RoBERTa, Mental-XLNet, and Mental-LongFormer \cite{ji-etal-2022-mentalbert, ji2023domain}, along with large generative language models (LLMs) like LLama-2 \cite{touvron2023llama}, GPT-3.5-turbo \cite{ouyang2022-instuctGPT}, and GPT-4 \cite{achiam2023gpt}. Our results unveil intriguing findings. While GPT-4 emerges as the top performer overall, the pretrained BERT-style models fine-tuned on \Silverdataset\ are formidable competitors. Notably, PLMs almost always outperform GPT-3.5-turbo in binary classification setups and Llama-2 in both comorbidity and binary classification tasks. However, none of the models achieve an F1 score exceeding $72\%$ in the comorbid multi-class classification or surpass $69\%$ and $75\%$ in depression and anxiety binary classification, respectively. These insights prompt reconsideration on the suitability of current language models for sensitive and nuanced tasks like mental health diagnosis \cite{de2023benefits, timmons2023call}.

\paragraph{Contributions.}
In summary, we make the following contributions: \footnote{The source code and dataset of this study are available at \href{https://github.com/AmeyHengle/ANGST}{https://github.com/AmeyHengle/ANGST}}
\begin{itemize}[leftmargin=*,noitemsep,topsep=0.5pt]
    \item \textbf{Novel Dataset} -- We curate \dataset, a meticulously crafted, gold-labeled, and neutrally-seeded Reddit corpus tailored for depression-anxiety comorbidity classification. \dataset\ can be employed in both multi-label and binary classification setups. Section \ref{sec:data} details the data sourcing, filtering, and annotation processes.
    \item \textbf{Silver Labels} -- We compile \Silverdataset, a silver-labeled dataset, which can complement \dataset\ for few-shot learning or supervised fine-tuning (Section \ref{sec:silver_data}).
    \item \textbf{Cross-dataset Analysis} -- We conduct a comprehensive examination of \dataset\ across various facets in comparison to existing mental health corpora, highlighting its unique characteristics (Section \ref{sec:data_analysis}).
    \item \textbf{Large-Scale Benchmarking} -- We evaluate numerous pretrained language models (PLMs) and large language models (LLMs) on \dataset, delineating their respective advantages and shortcomings (Section \ref{sec:results_and_discussion}).
    \item \textbf{Qualitative and Quantitative Evaluation} -- We thoroughly analyze the results across different metrics, complemented by an in-depth error analysis, providing a holistic assessment of model performance (Section \ref{sec:error_analysis}).
\end{itemize}

\begin{table*}[!ht]
\begin{center}
\resizebox{\textwidth}{!}{%
\begin{tabular}{p{7.2in}cc}
\toprule
\textbf{Example} & \textbf{Depression} & \textbf{Anxiety} \\
\midrule
\textit{I no longer hate myself in the same way. I feel somewhat better now that I've accepted my ugly appearance, shy personality, and few mental health issues. Before, I used to despise myself for these things. Although I am aware that I will never be loved, have a successful career, be a good friend, or find a girlfriend, I no longer feel horrible about myself because I accept how boring and ugly I am. The only thing that worries me is that I am more foolish than I care to acknowledge.} & \textcolor{red}{\ding{55}} & \textcolor{red}{\ding{55}} \\
\cmidrule(lr){1-3}
\textit{I'm in pain all the time. Really, what's the purpose of existence? 
\textcolor{red}{I literally experience daily, unrelenting emotional pain.} and the best i can do is distract myself from it. \textcolor{red}{I literally experience constant emotional pain every single day. Most of the time, all I want to do is curl up into a ball and cry my eyes out, but I'm so disconnected that I can't cry anymore.}. I mean, I won't kill myself, but when every day is agony, how is it possible for anyone to really think that I would want to live? Nothing ever seems to improve the situation, and I can feel my condition gradually deteriorating every day. \textcolor{red}{I wish I could just vanish.}
} & \textcolor{ForestGreen}{\ding{51}} & \textcolor{red}{\ding{55}} \\
\cmidrule(lr){1-3}
\textit{What just I experienced. I suddenly experienced a wave of extreme anxiety around 9:40 p.m. I had to lock every door, but I swear I saw someone when I peered outside. I swear I saw someone, but they would have to be taller than seven feet. My dad hurried to tell my parents after he went outside and saw nobody. This should normally calm people down, right? \textcolor{blue}{It didnt ease me}. \textcolor{blue}{I still feel as though I'm going to die or am in danger.} \textcolor{blue}{I couldnt stop shaking} and I was only able to stop crying at 10 p.m. \textcolor{blue}{My chest and head hurt really bad now}. Was this an extreme case of anxiety, or something else?
}& \textcolor{red}{\ding{55}} & \textcolor{ForestGreen}{\ding{51}} \\
\cmidrule(lr){1-3}
\textit{Stuck after graduation. August was when I graduated, and I still haven't found employment in my field. I continue to work a few hours a week at my menial retail job, and because I am so \textcolor{red}{severely depressed and lonely}, I am afraid to apply to as many jobs as I should because \textcolor{blue}{I feel incompetent and afraid, literally frozen with fear.} I'll die over it. I spend everyday in a \textcolor{red}{constant cycle of loneliness} and \textcolor{blue}{anxiety and worry and panic and emptiness}. I feel like I'm burdening my family more and more every day because I can't seem to get a job and get my life started. I simply want to put an end to this years-long suffering because it keeps getting worse. I despise myself for it, \textcolor{red}{Due to family issues, I've literally cut off from my friends and am dreading Christmas.} and talk about why Im not working yet.}& \textcolor{ForestGreen}{\ding{51}} & \textcolor{ForestGreen}{\ding{51}} \\
\bottomrule
\end{tabular}%
}
\end{center}\caption{Examples of depression, anxiety, comorbidity, and control posts in \dataset. \textcolor{red}{Red} signifies signs of depression and \textcolor{blue}{Blue} highlights anxiety symptoms. All of the posts in this example have been paraphrased in order to protect the user's identity.}
\label{tab:examples}
\end{table*}
\section{Related Work}
\paragraph{Corpora related to Mental Health Disorders.} 
Since the past decade, social media platforms have been actively used for compiling datasets for various mental health issues \cite{de-choudhary-etal-2013, de2013predicting, tsugawa-etal-2015, pedersen-2015-screening}. For instance, CLPsych15 \cite{coppersmith-etal-2015-clpsych}, RSDD \cite{yates-etal-2017-depression}, Depression Reddit \cite{pirina-coltekin-2018-identifying}, and Dreaddit \cite{turcan-mckeown-2019-dreaddit} are commonly used corpora for depression analysis. Similarly, DATD \cite{owen-etal-2020-towards} and the Anxiety on Reddit corpus \cite{shen-rudzicz-2017-detecting} are popular benchmarks for anxiety issues. UMD \cite{shing-etal-2018-expert} and T-SID \cite{ji2020suicidalb} are compiled exclusively for identifying suicide ideation. 

The CLPsych16 dataset \cite{milne-etal-2016-clpsych} supports multi-label classification, covering depression, control, and PTSD categories. Additionally, the CAMS \cite{garg-etal-2022-cams} and SAD \cite{SAD-dataset} datasets offer more granular classifications, categorizing the causes of depression into six and nine categories, respectively.
On a broader scale, SMHD \cite{cohan-etal-2018-smhd}, SWMH \cite{ji2022suicidala}, and CAMS \cite{garg-etal-2022-cams} encompass a more comprehensive coverage of several mental health disorders. \citet{yang2023mentalllama} aggregated multiple mental health-related datasets into a unified benchmark named IMHI.  More recently, \citet{jin2023psyeval} proposed PsyEval, a suite of mental health-related tasks designed to evaluate the performance of large language models.

\paragraph{Computational Models for Mental Health Disorder Identification. } Early approaches for identifying mental health disorders from social media platforms relied heavily on feature engineering and traditional machine learning classifiers \cite{de-choudhary-etal-2013, de2013predicting, Coppersmith_Harman_Dredze_2014, mitchell-etal-2015-quantifying, tsugawa-etal-2015}. Subsequent research endeavors aimed to obviate the need for hand-crafted features by employing neural network architectures such as LSTMs and CNNs, enabling more accurate identification of mental health conditions \cite{sawhney-etal-2018-exploring, tadesse2019detection}. 
\citet{de-souza-etal-2022} propose a deep learning ensemble method that effectively classifies anxiety, depression, and their comorbidity using Reddit posts.
Another line of research explores the use of mental health questionnaires \cite{nguyen-etal-2022-improving}, multi-task learning \cite{sarkar2022predicting}, and hierarchical attention networks \cite{han-etal-2022-hierarchical} to augment existing architectures. 
In recent years, numerous transformer-based models pre-trained on mental health-related data, such as Mental-BERT, Mental-RoBERTa, Mental-XLNet, and Mental-LongFormer \cite{ji-etal-2022-mentalbert, ji2023domain}, have been developed and released. More recently, several open-source mental health-focused LLMs, including  Mental-Flant-T5, Mental-Alpaca \cite{xu2024mental}, and Mental-LLama \cite{yang2023mentalllama} have been introduced. Furthermore, endeavors such as that of \citet{chen2023chatgpt}, \citet{yang2023evaluations}, and \citet{yang-etal-2023-towards} provide in-depth analyses of proprietary LLMs like chatGPT, GPT-3, and GPT-4, evaluating their performance in identifying various mental health conditions.

\section{Curating \dataset}
\label{sec:data}

\subsection{Data Collection and Filtering}

\dataset\ is compiled from publicly available Reddit posts. As illustrated in Table \ref{tab:examples},  each post in \dataset\ is labeled in a multi-label fashion to indicate depression and/or anxiety. We adhere to established protocols for data collection as described in prior research on constructing mental health datasets \cite{yates-etal-2017-depression, cohan-etal-2018-smhd}. To begin with, we compile a list of relevant mental-health-related Subreddits informed by prior investigations into depression and anxiety on Reddit \cite{pavalanathan-choudhary-2015, yates-etal-2017-depression, cohan-etal-2018-smhd}. We employ PRAW ( Python Reddit API Wrapper) \footnote{\url{https://praw.readthedocs.io/en/stable}} to scrape all publicly available Reddit posts in these Subreddits over five years from January 2018 to December 2022. This resulted in approximately $400,000$ anonymous posts. 

Next, as outlined by \citet{yates-etal-2017-depression}, we identified authors who had self-disclosed their experiences with depression, anxiety, or related disorders using high-precision diagnostic patterns \cite{cohan-etal-2018-smhd}. We identified $8,083$ unique authors who had made at least one self-disclosure post. Subsequently, we compiled all posts made by these authors, resulting in a corpus of $185,064$ posts. Further refining this dataset, we excluded posts shorter than $75$ words and those authored by individuals with fewer than $10$ posts, producing a more concentrated corpus of $72,583$ posts. From this filtered corpus, we selected the top $25,000$ posts that exhibited the highest linguistic alignment with depression and anxiety cues. Specifically, we identified and ranked posts based on their Empath \cite{empath-2016} and NRC \cite{saif-mohammad-nrc-1,saif-mohammad-nrc-2} sentiment scores. We use the Empath and NRC scores to prioritize posts that were more likely to contain pertinent emotional and psychological indicators associated with depression and/or anxiety, in light of prior research that highlights their effectiveness \cite{hui-etal-2022-liwc, marco-etal-2023-liwc, zhang-2023-liwc-empath-nrc}. Appendix Table \ref{tab:nrc_empath} provides examples of posts with low, moderate, and high Empath and NRC scores, respectively. As illustrated in the table, posts with higher scores show a stronger alignment with the linguistic cues of depression and anxiety. Finally, we randomly sample $3000$ posts from this pool to form our final annotation corpus.

\subsection{Data Annotation}
We employed two primary and one secondary annotator, each a trained professional psychologist. To preserve objectivity and prevent potential biases, we established a protocol ensuring complete anonymity between the annotators. Specifically, no additional information or context beyond the post's content was shared with them. Moreover, the annotators worked in isolation, ensuring their annotations were performed independently, precluding any collaborative influence. The primary role of the annotators was twofold. First, they discerned and labeled if a post signified depression or anxiety, indicating their decision with a clear "yes" or "no." Second, highlight specific statements in the posts to support the validity of their judgment (see Table \ref{tab:examples}). The secondary annotator is used to resolve conflicting annotations. Additionally, to protect the identity of users, the annotators were asked to flag any posts that might inadvertently disclose user details, including names, demographics, or other personal identifiers. This resulted in some posts being flagged, and the final remaining corpus consisted of $2,876$ posts, forming our \dataset\ dataset.

We report the inter-annotator agreement between our primary annotators for both depression and anxiety labels using  Krippendorff's alpha ($\alpha$) \cite{Krippendorff2011ComputingKA}, and Fleiss kappa ($\kappa$) \cite{cohen-kappa}. For depression, the agreement scores are $\alpha = 0.622$ and $\kappa = 0.624$, while for anxiety, the scores are $\alpha = 0.423$ and $\kappa = 0.444$. These metrics indicate a moderate level of agreement, within the acceptable range of $0.4$ to $0.7$. Notably, the agreement on depression labels is higher compared to anxiety, reflecting a more consistent identification of depressive symptoms among the annotators.
\section{Yet Another Mental Health Dataset?}
\label{sec:data_analysis}

In this section, we conduct a comprehensive cross-sectional analysis of \dataset. We scrutinize it across various facets and juxtapose it with existing mental health corpora, thereby underlining its unique characteristics and distinguishing features. Specifically, we compare \dataset\ against SDCNL \cite{SDCNL-dataset}, Depression Reddit \cite{pirina-coltekin-2018-identifying} Dreaddit \cite{turcan-mckeown-2019-dreaddit}, and DATD \cite{owen-etal-2020-towards}. We employ the corpora comparison strategy as outlined by \citet{kulkarni2023revisiting}.

\subsection{Inter-Class Similarity}

\begin{table}[t!]
\begin{center}
\small 
\setlength{\tabcolsep}{4pt} 
\resizebox{0.95\columnwidth}{!}{
\begin{tabular}{lllcccc}
\toprule
\multirow{2}{*}{\textbf{Dataset}} & \multirow{2}{*}{\textbf{Label 1}} & \multirow{2}{*}{\textbf{Label 2}} &  \multicolumn{2}{c}{\textbf{JSD}} &  \multicolumn{2}{c}{\textbf{MMD}}\\
\cmidrule(lr){4-5}
\cmidrule(lr){6-7}
& & & \textbf{Unigram} & \textbf{Bigram} & \textbf{Mean} & \textbf{Median} \\
\midrule
\multirow{6}{*}{\dataset} & control & comorbid & $0.036$ & $0.154$ & $0.047$ & $0.043$ \\
& control & depression & $0.027$ & $0.142$ & $0.047$ & $0.043$ \\
& control & anxiety & $0.034$& $0.150$ & $0.048$& $0.045$ \\
& depression & anxiety & $0.014$ & $0.070$ & $0.041$ & $0.038$ \\
& depression & comorbid & $0.002$ & $0.011$ & $0.042$ & $0.038$ \\
& anxiety & comorbid & $0.013$ & $0.068$ & $0.041$ & $0.038$ \\
\cmidrule(lr){1-7}
DATD & control & anxiety & $0.104$ & $0.182$ & $0.083$ & $0.056$\\
\cmidrule(lr){1-7}
Dep Reddit & control & depression & $0.060$ & $0.165$ & $0.082$ & $0.073$ \\
\cmidrule(lr){1-7}
Dreaddit & control & Stress & $0.056$ & $0.174$  & $0.067$ & $0.063$ \\
\cmidrule(lr){1-7}
SDCNL & depression & Suicide & $0.031$ & $0.150$ & $0.070$ & $0.060$ \\
\bottomrule
\end{tabular}%
}
\end{center}
\caption{Inter-class similarity in different mental health datasets measured via Jensen–Shannon Divergence (JSD) and Maximum Mean Discrepancy (MMD). The lower the values of these metrics, the harder it will be to separate the classes.}
\label{tab:jsd}
\end{table}

\textbf{Hypothesis} -- The more similar the inter-class samples, the more difficult it becomes to classify the dataset. We hypothesize that due to its neutral-seeding policy, \dataset\ would exhibit higher inter-class similarity than its counterparts, thereby rendering it more challenging.

\paragraph{Metrics} -- We utilize the Jensen–Shannon divergence (JSD) \cite{dagan-etal-1997-similarity} and Maximum Mean Discrepancy (MMD) \cite{gretton12a} as measures to quantify the proximity between different classes in a dataset. To calculate MMD, we use RoBERTa embeddings \cite{liu2019roberta}. The transformer weights were frozen, and a classification head with the required number of class labels was employed for each experiment. A lower value of JSD or MMD indicates that the class distributions are more similar, making it more challenging to classify accurately.

\paragraph{Experiments} -- we generate a Laplacian smoothed unigram and bigram distribution for each dataset for each class. Table \ref{tab:jsd} presents the Jensen–Shannon divergence (JSD) of the pairwise selection of labels. For \dataset, the JSD of depression, anxiety, and comorbidity versus the control group are $0.027$, $0.034$, and $0.036$, respectively, suggesting that the comorbid and control examples exhibit the least similarity. The JSD of anxiety with depression and comorbidity is $0.014$ and $0.013$, while that of depression and comorbidity is $0.002$. This low divergence is what renders the dataset arduous to classify and is the cause for the high disagreement in the non-control classes.

When extending the above experiment with other datasets, we observe that the pairwise JSD of \dataset\ is lower than that of DATD ($0.104$), Depression Reddit ($0.060$), and Dreaddit ($0.056$). The higher divergence in these datasets is expected, as their control group samples are sourced from non-mental health contexts. Thus, \dataset\ serves as a superior benchmark, as it captures the vital yet minute differences between anxiety and depression more effectively. Results for MMD are described in Section \ref{sec:mmd}.

\subsection{Adversarial Validation}
\textbf{Hypothesis} -- Data drift \cite{lu2018learning} quantifies the change in the feature space between two datasets. All samples in the old (source) dataset are considered the negative class, and all the samples in the new (target) dataset are deemed the positive class. A simple classifier is trained for this binary classification task. A high performance suggests the presence of discriminatory features between the two datasets. We conjecture that due to its meticulous data curation and filtering pipeline and its gold annotation scheme, \dataset\ will exhibit significant differences compared to existing mental health datasets.

\paragraph{Metrics} -- We analyze the performance based on accuracy, macro-F1, ROC-AUC scores, and Matthews Correlation Coefficient.

\paragraph{Experiment} -- The training split ($T$) from the source dataset $X^{(T)}_{\text{src}}$ is labeled as ($X^{(T)}_{\text{src}} = 0$), and \dataset's training split $X^{(T)}_{\text{tgt}}$ is labeled as ($X^{(T)}_{\text{tgt}} = 1$). We utilize an n-gram-based TF-IDF logistic regression model to detect the data drift of \dataset\ concerning the chosen datasets. As shown in Table \ref{tab:adv}, none of the dataset comparisons result in a ROC-AUC less than $0.5$ or an MCC approximately equal to $0$, which would have suggested that our dataset is indistinguishable from existing datasets. On the contrary, the results demonstrate variation within a narrow range of $0.096$ for ROC-AUC and $0.125$ for MCC. The highest and lowest ROC-AUC scores are 1.0 and 0.904, obtained from SDCNL and DATD, respectively, while the corresponding MCC scores are 0.990 and 0.875. This indicates that \dataset\ exhibits an inherent difference from existing corpora, which can be attributed to its meticulous data curation, filtering pipeline, and gold annotation scheme.

\begin{table*}[!t!]
\begin{center}
\resizebox{\textwidth}{!}{%
\begin{tabular}{lccccccccccc}
\toprule
\multirow{2}{*}{\textbf{Model}} & \multirow{2}{*}{\textbf{Type}} & 
\multicolumn{5}{c}{\textbf{Depression vs. Control}} & 
\multicolumn{5}{c}{\textbf{Anxiety vs. Control}}\\
\cmidrule(lr){3-7} 
\cmidrule(lr){8-12} 
 &  & \textbf{Acc} & \textbf{Precision} & \textbf{Recall} & \underline{\textbf{F1}} & \textbf{macro-F1} & \textbf{Acc} & \textbf{Precision} & \textbf{Recall} & \underline{\textbf{F1}} & \textbf{macro F1} \\
\midrule
Mental-BERT$_{\mathrm{base}}^\ddag$ & SFT & \cellcolor{yellow!25}\textcolor{black}{$68.6_{1.7}$} & \cellcolor{yellow!40}\textcolor{black}{\underline{$\boldsymbol{77.5_{1.3}}$}} & \cellcolor{yellow!25}\textcolor{black}{$68.6_{1.7}$} & \cellcolor{yellow!15}\textcolor{black}{$64.3_{2.6}$} & \cellcolor{yellow!10}\textcolor{black}{$62.8_{2.7}$} & \cellcolor{yellow!45}\textcolor{black}{\underline{$\boldsymbol{78.3_{1.1}}$}} & \cellcolor{yellow!50}\textcolor{black}{$79.9_{1.8}$} & \cellcolor{yellow!45}\textcolor{black}{\underline{$\boldsymbol{78.8_{1.1}}$}} & \cellcolor{yellow!45}\textcolor{black}{$78.3_{1.3}$} & \cellcolor{yellow!35}\textcolor{black}{$72.2_{2.1}$}\\
Mental-BERT$_{\mathrm{large}}$ & SFT & \cellcolor{yellow!25}\textcolor{black}{$68.7_{1.9}$} & \cellcolor{yellow!35}\textcolor{black}{$76.6_{0.9}$} & \cellcolor{yellow!25}\textcolor{black}{$68.7_{1.9}$} & \cellcolor{yellow!15}\textcolor{black}{$64.6_{3.0}$} & \cellcolor{yellow!10}\textcolor{black}{$63.1_{3.3}$} & \cellcolor{yellow!35}\textcolor{black}{$76.5_{1}$} & \cellcolor{yellow!40}\textcolor{black}{$77.6_{1.8}$} & \cellcolor{yellow!35}\textcolor{black}{$76.5_{1.0}$} & \cellcolor{yellow!35}\textcolor{black}{$76.9_{1.2}$} & \cellcolor{yellow!20}\textcolor{black}{$69.2_{2.2}$} \\
Mental-RoBERTa$_{\mathrm{large}}$ & SFT & \cellcolor{yellow!25}\textcolor{black}{$68.9_{1.5}$} & \cellcolor{yellow!35}\textcolor{black}{$76.4_{1.3}$} & \cellcolor{yellow!25}\textcolor{black}{$68.9_{1.5}$} & \cellcolor{yellow!15}\textcolor{black}{$65.0_{2.3}$} & \cellcolor{yellow!10}\textcolor{black}{$63.6_{2.4}$} & \cellcolor{yellow!35}\textcolor{black}{$76.2_{0.9}$} & \cellcolor{yellow!40}\textcolor{black}{$77.7_{1.8}$} & \cellcolor{yellow!35}\textcolor{black}{$76.2_{0.9}$} & \cellcolor{yellow!35}\textcolor{black}{$76.7_{1.1}$} & \cellcolor{yellow!20}\textcolor{black}{$69.3_{2.2}$} \\
Mental-XLNet$_{\mathrm{base}}^\ddag$ & SFT & \cellcolor{yellow!30}\textcolor{black}{$69.8_{1.0}$} & \cellcolor{yellow!35}\textcolor{black}{$76.9_{1.5}$} & \cellcolor{yellow!30}\textcolor{black}{$69.8_{1.0}$} & \cellcolor{yellow!20}\textcolor{black}{$66.3_{1.5}$} & \cellcolor{yellow!15}\textcolor{black}{$65.0_{1.6}$} & \cellcolor{yellow!35}\textcolor{black}{$76.9_{0.8}$} & \cellcolor{yellow!40}\textcolor{black}{$78.2_{1.6}$} & \cellcolor{yellow!35}\textcolor{black}{$76.9_{0.8}$} & \cellcolor{yellow!40}\textcolor{black}{$77.4_{1.0}$} & \cellcolor{yellow!25}\textcolor{black}{$70.0_{1.9}$} \\
Mental-LongFormer$_{\mathrm{base}}$ & SFT & \cellcolor{yellow!25}\textcolor{black}{$68.7_{1.1}$} & \cellcolor{yellow!35}\textcolor{black}{$76.3_{1.1}$} & \cellcolor{yellow!25}\textcolor{black}{$68.7_{1.1}$} & \cellcolor{yellow!15}\textcolor{black}{$64.7_{1.7}$} & \cellcolor{yellow!10}\textcolor{black}{$63.3_{1.8}$} & \cellcolor{yellow!40}\textcolor{black}{$77.5_{1.1}$} & \cellcolor{yellow!45}\textcolor{black}{$78.8_{2.0}$} & \cellcolor{yellow!40}\textcolor{black}{$77.5_{1.1}$} & \cellcolor{yellow!45}\textcolor{black}{$77.9_{1.3}$} & \cellcolor{yellow!25}\textcolor{black}{$70.8_{2.2}$} \\
\midrule
\multirow{2}{*}{LLama-2-chat-7B} & zero-shot & \cellcolor{yellow!10}\textcolor{black}{$55.7_{0.5}$} & \cellcolor{yellow!15}\textcolor{black}{$57.3_{2.2}$} & \cellcolor{yellow!10}\textcolor{black}{$55.7_{0.5}$} & \cellcolor{yellow!5}\textcolor{black}{$43_{1.2}$} & \cellcolor{yellow!5}\textcolor{black}{$39.9_{1.3}$} & \cellcolor{yellow!5}\textcolor{black}{$25.4_{0.4}$} & \cellcolor{yellow!30}\textcolor{black}{$66.2_{3.5}$} & \cellcolor{yellow!5}\textcolor{black}{$25.4_{0.4}$} & \cellcolor{yellow!5}\textcolor{black}{$12_{0.9}$} & \cellcolor{yellow!5}\textcolor{black}{$21.2_{0.6}$}\\
& few-shot & \cellcolor{yellow!5}\textcolor{black}{$51.6_{0.1}$} & \cellcolor{yellow!5}\textcolor{black}{$52_{0.1}$} & \cellcolor{yellow!5}\textcolor{black}{$51.6_{0.1}$} & \cellcolor{yellow!5}\textcolor{black}{$51.7_{0.1}$} & \cellcolor{yellow!5}\textcolor{black}{$51.5_{0.1}$} & \cellcolor{yellow!10}\textcolor{black}{$43.8_{1}$} & \cellcolor{yellow!30}\textcolor{black}{$64.1_{0.8}$} & \cellcolor{yellow!10}\textcolor{black}{$43.8_{1}$} & \cellcolor{yellow!15}\textcolor{black}{$46.5_{1.1}$} & \cellcolor{yellow!10}\textcolor{black}{$43_{0.9}$}\\ 
\cmidrule(lr){1-12}
\multirow{2}{*}{LLama-2-chat-13B} & zero-shot & \cellcolor{yellow!15}\textcolor{black}{$60.9_{0.2}$} & \cellcolor{yellow!20}\textcolor{black}{$62.5_{0.2}$} & \cellcolor{yellow!15}\textcolor{black}{$60.9_{0.2}$} & \cellcolor{yellow!10}\textcolor{black}{$56.6_{0.4}$} & \cellcolor{yellow!10}\textcolor{black}{$54.9_{0.4}$} & \cellcolor{yellow!5}\textcolor{black}{$39.8_{0.5}$} & \cellcolor{yellow!30}\textcolor{black}{$68.1_{0.3}$} & \cellcolor{yellow!5}\textcolor{black}{$39.8_{0.5}$} & \cellcolor{yellow!5}\textcolor{black}{$39.8_{0.7}$} & \cellcolor{yellow!5}\textcolor{black}{$39.8_{0.5}$} \\
& few-shot & \cellcolor{yellow!5}\textcolor{black}{$46.2_{0.9}$} & \cellcolor{yellow!10}\textcolor{black}{$49.1_{1.5}$} & \cellcolor{yellow!5}\textcolor{black}{$46.2_{0.9}$} & \cellcolor{yellow!5}\textcolor{black}{$41.1_{1}$} & \cellcolor{yellow!5}\textcolor{black}{$42.6_{0.9}$} & \cellcolor{yellow!15}\textcolor{black}{$54.9_{1}$} & \cellcolor{yellow!25}\textcolor{black}{$62.7_{0.8}$} & \cellcolor{yellow!15}\textcolor{black}{$54.9_{1}$} & \cellcolor{yellow!15}\textcolor{black}{$57.7_{0.9}$} & \cellcolor{yellow!15}\textcolor{black}{$48.2_{1}$} \\
\cmidrule(lr){1-12}
\multirow{2}{*}{GPT-3.5-turbo} & zero-shot & \cellcolor{yellow!20}\textcolor{black}{$64.1_{0.1}$} & \cellcolor{yellow!35}\textcolor{black}{$71.3_{0.1}$} & \cellcolor{yellow!20}\textcolor{black}{$64.1_{0.1}$} & \cellcolor{yellow!15}\textcolor{black}{$58.3_{0.1}$} & \cellcolor{yellow!15}\textcolor{black}{$56.5_{0.1}$} & \cellcolor{yellow!5}\textcolor{black}{$41.2_{0.1}$} & \cellcolor{yellow!45}\textcolor{black}{$77.3_{0.1}$} & \cellcolor{yellow!5}\textcolor{black}{$41.2_{0.1}$} & \cellcolor{yellow!5}\textcolor{black}{$39.6_{0.1}$} & \cellcolor{yellow!5}\textcolor{black}{$41.0_{0.1}$} \\
& few-shot & \cellcolor{yellow!20}\textcolor{black}{$64.8_{0.0}$} & \cellcolor{yellow!35}\textcolor{black}{$70.1_{0.1}$} & \cellcolor{yellow!20}\textcolor{black}{$64.8_{0.0}$} & \cellcolor{yellow!20}\textcolor{black}{$60.2_{0.0}$} & \cellcolor{yellow!15}\textcolor{black}{$58.6_{0.0}$} & \cellcolor{yellow!5}\textcolor{black}{$38.4_{0.1}$} & \cellcolor{yellow!45}\textcolor{black}{$77.2_{0.1}$} & \cellcolor{yellow!5}\textcolor{black}{$38.4_{0.1}$} & \cellcolor{yellow!5}\textcolor{black}{$35.4_{0.2}$} & \cellcolor{yellow!5}\textcolor{black}{$38.0_{0.2}$} \\
\cmidrule(lr){1-12}
\multirow{2}{*}{GPT-4} & zero-shot$^\dagger$ & \cellcolor{yellow!45}\textcolor{black}{\underline{$\boldsymbol{70.7_{0.0}}$}} & \cellcolor{yellow!50}\textcolor{black}{$74.7_{0.2}$} & \cellcolor{yellow!45}\textcolor{black}{\underline{$\boldsymbol{70.7_{0.0}}$}} & \cellcolor{yellow!40}\textcolor{black}{\underline{$\boldsymbol{68.4_{0.1}}$}} & \cellcolor{yellow!40}\textcolor{black}{\underline{$\boldsymbol{67.4_{0.1}}$}} & \cellcolor{yellow!45}\textcolor{black}{$77.6_{0.1}$} & \cellcolor{yellow!50}\textcolor{black}{\underline{$\boldsymbol{83.3_{0.0}}$}} & \cellcolor{yellow!45}\textcolor{black}{$77.6_{0.1}$} & \cellcolor{yellow!45}\textcolor{black}{\underline{$\boldsymbol{78.9_{0.1}}$}} & \cellcolor{yellow!40}\textcolor{black}{\underline{$\boldsymbol{74.0_{0.1}}$}} \\
& few-shot & \cellcolor{yellow!25}\textcolor{black}{$65.5_{0.1}$} & \cellcolor{yellow!40}\textcolor{black}{$76.7_{0.1}$} & \cellcolor{yellow!25}\textcolor{black}{$65.5_{0.1}$} & \cellcolor{yellow!20}\textcolor{black}{$59.4_{0.1}$} & \cellcolor{yellow!15}\textcolor{black}{$57.6_{0.1}$} & \cellcolor{yellow!30}\textcolor{black}{$68.1_{0.2}$} & \cellcolor{yellow!45}\textcolor{black}{$82.5_{0.1}$} & \cellcolor{yellow!30}\textcolor{black}{$68.1_{0.2}$} & \cellcolor{yellow!35}\textcolor{black}{$70.3_{0.2}$} & \cellcolor{yellow!25}\textcolor{black}{$66.2_{0.2}$} \\
\midrule
\multicolumn{2}{c}{$\Delta_{\mathrm{model}^\dagger - \mathrm{model}^\ast}$} & \textcolor{ForestGreen}{$\uparrow0.9$} & \textcolor{red}{$\downarrow2.2$} & \textcolor{ForestGreen}{$\uparrow0.9$} & \textcolor{ForestGreen}{$\uparrow2.1$} & \textcolor{ForestGreen}{$\uparrow2.4$} & \textcolor{red}{$\downarrow0.$} & \textcolor{ForestGreen}{$\uparrow3.4$} & \textcolor{red}{$\downarrow1.2$} & \textcolor{ForestGreen}{$\uparrow0.6$} & \textcolor{ForestGreen}{$\uparrow1.8$} \\
\bottomrule
\end{tabular}%
}
\end{center}\caption{Results of various methods for depression and anxiety binary classification.}
\label{tab:results_binary}
\end{table*}

\begin{table}[!ht]
\begin{center}
\resizebox{\columnwidth}{!}{%
\begin{tabular}{lccccccc}
\toprule
\multirow{2}{*}{\textbf{Model}} & \multirow{2}{*}{\textbf{Type}} & 
\multicolumn{3}{c}{\textbf{Depression}} & 
\multicolumn{3}{c}{\textbf{Control}}\\
\cmidrule(lr){3-5}
\cmidrule(lr){6-8}
 &  & \textbf{Precision} & \textbf{Recall} & \textbf{F1} & \textbf{Precision} & \textbf{Recall} & \textbf{F1} \\
\midrule
Mental-BERT$_{\mathrm{base}}$ & SFT & \cellcolor{blue!25}\textcolor{black}{$64.0_{1.3}$} & \cellcolor{blue!50}\textcolor{black}{$98.2_{0.8}$} & \cellcolor{blue!40}\textcolor{black}{$77.5_{0.9}$} & \cellcolor{blue!50}\textcolor{black}{$94.0_{2.5}$} & \cellcolor{blue!10}\textcolor{black}{$32.5_{4.2}$} & \cellcolor{blue!15}\textcolor{black}{$48.1_{4.6}$}\\
Mental-BERT$_{\mathrm{large}}$ & SFT & \cellcolor{blue!25}\textcolor{black}{$64.3_{1.7}$} & \cellcolor{blue!45}\textcolor{black}{$97.3_{1.6}$} & \cellcolor{blue!40}\textcolor{black}{$77.4_{0.8}$} & \cellcolor{blue!45}\textcolor{black}{$91.7_{3.2}$} & \cellcolor{blue!10}\textcolor{black}{$33.7_{5.7}$} & \cellcolor{blue!15}\textcolor{black}{$48.9_{5.8}$}\\
Mental-RoBERTa$_{\mathrm{large}}$ & SFT & \cellcolor{blue!25}\textcolor{black}{$64.4_{1.3}$} & \cellcolor{blue!45}\textcolor{black}{$97.2_{1.2}$} & \cellcolor{blue!40}\textcolor{black}{$77.5_{0.8}$} & \cellcolor{blue!40}\textcolor{black}{$91.1_{2.9}$} & \cellcolor{blue!10}\textcolor{black}{$34.4_{4.1}$} & \cellcolor{blue!20}\textcolor{black}{$49.7_{4.2}$}\\
Mental-XLNet$_{\mathrm{base}}^\ast$ & SFT & \cellcolor{blue!30}\textcolor{black}{$65.2_{0.9}$} & \cellcolor{blue!45}\textcolor{black}{$97.0_{1.7}$} & \cellcolor{blue!40}\textcolor{black}{$77.9_{0.6}$} & \cellcolor{blue!40}\textcolor{black}{$91.2_{3.6}$} & \cellcolor{blue!15}\textcolor{black}{$36.7_{3.3}$} & \cellcolor{blue!25}\textcolor{black}{$52.2_{2.9}$}\\
Mental-LongFormer$_{\mathrm{base}}$ & SFT & \cellcolor{blue!25}\textcolor{black}{$64.3_{0.9}$} & \cellcolor{blue!45}\textcolor{black}{$97.2_{1.0}$} & \cellcolor{blue!40}\textcolor{black}{$77.4_{0.6}$} & \cellcolor{blue!40}\textcolor{black}{$91.0_{2.7}$} & \cellcolor{blue!10}\textcolor{black}{$33.9_{3.1}$} & \cellcolor{blue!20}\textcolor{black}{$49.3_{3.2}$} \\
\midrule
\multirow{2}{*}{LLama-2-chat-7B} & zero-shot & \cellcolor{blue!10}\textcolor{black}{$55.6_{0.3}$} & \cellcolor{blue!45}\textcolor{black}{$97.3_{0.2}$} & \cellcolor{blue!30}\textcolor{black}{$70.7_{0.2}$} & \cellcolor{blue!5}\textcolor{black}{$59.3_{4.4}$} & \cellcolor{blue!5}\textcolor{black}{$5_{1.4}$} & \cellcolor{blue!5}\textcolor{black}{$9.2_{2.4}$} \\
& few-shot & \cellcolor{blue!10}\textcolor{black}{$56.5_{0.1}$} & \cellcolor{blue!5}\textcolor{black}{$52.4_{0.5}$} & \cellcolor{blue!10}\textcolor{black}{$54.3_{0.3}$} & \cellcolor{blue!5}\textcolor{black}{$46.6_{0.1}$} & \cellcolor{blue!5}\textcolor{black}{$50.7_{0.5}$} & \cellcolor{blue!10}\textcolor{black}{$48.6_{0.2}$} \\
\cmidrule(lr){1-8}
\multirow{2}{*}{LLama-2-chat-13B} & zero-shot & \cellcolor{blue!15}\textcolor{black}{$59.7_{0.2}$} & \cellcolor{blue!40}\textcolor{black}{$88.4_{0.4}$} & \cellcolor{blue!30}\textcolor{black}{$71.3_{0.1}$} & \cellcolor{blue!15}\textcolor{black}{$65.8_{0.3}$} & \cellcolor{blue!5}\textcolor{black}{$27.2_{0.9}$} & \cellcolor{blue!10}\textcolor{black}{$38.5_{0.9}$} \\
& few-shot & \cellcolor{blue!5}\textcolor{black}{$52.9_{2.3}$} & \cellcolor{blue!5}\textcolor{black}{$19.1_{0.8}$} & \cellcolor{blue!5}\textcolor{black}{$28.1_{1.2}$} & \cellcolor{blue!10}\textcolor{black}{$44.5_{0.6}$} & \cellcolor{blue!45}\textcolor{black}{$79.2_{1.1}$} & \cellcolor{blue!40}\textcolor{black}{$57_{0.7}$} \\
\cmidrule(lr){1-8}
\multirow{2}{*}{GPT-3.5-turbo} & zero-shot & \cellcolor{blue!20}\textcolor{black}{$61.0_{0.0}$} & \cellcolor{blue!50}\textcolor{black}{$96.1_{0.0}$} & \cellcolor{blue!40}\textcolor{black}{$74.6_{0.0}$} & \cellcolor{blue!50}\textcolor{black}{$84.0_{0.2}$} & \cellcolor{blue!10}\textcolor{black}{$24.9_{0.1}$} & \cellcolor{blue!10}\textcolor{black}{$38.4_{0.1}$} \\
& few-shot & \cellcolor{blue!20}\textcolor{black}{$61.8_{0.0}$} & \cellcolor{blue!45}\textcolor{black}{$94.2_{0.1}$} & \cellcolor{blue!40}\textcolor{black}{$74.6_{0.0}$} & \cellcolor{blue!45}\textcolor{black}{$80.3_{0.1}$} & \cellcolor{blue!10}\textcolor{black}{$28.9_{0.1}$} & \cellcolor{blue!15}\textcolor{black}{$42.5_{0.1}$} \\
\cmidrule(lr){1-8}
\multirow{2}{*}{GPT-4} & zero-shot$^\dagger$ & \cellcolor{blue!30}\textcolor{black}{\underline{$\boldsymbol{66.7_{0.1}}$}} & \cellcolor{blue!45}\textcolor{black}{$93.5_{0.3}$} & \cellcolor{blue!40}\textcolor{black}{\underline{$\boldsymbol{77.9_{0.0}}$}} & \cellcolor{blue!50}\textcolor{black}{$84.5_{0.4}$} & \cellcolor{blue!30}\textcolor{black}{\underline{$\boldsymbol{42.9_{0.4}}$}} & \cellcolor{blue!40}\textcolor{black}{\underline{$\boldsymbol{56.9_{0.2}}$}} \\
& few-shot & \cellcolor{blue!20}\textcolor{black}{$61.6_{0.0}$} & \cellcolor{blue!50}\textcolor{black}{\underline{$\boldsymbol{98.9_{0.0}}$}} & \cellcolor{blue!40}\textcolor{black}{$75.9_{0.0}$} & \cellcolor{blue!50}\textcolor{black}{\underline{$\boldsymbol{95.0_{0.2}}$}} & \cellcolor{blue!10}\textcolor{black}{$24.7_{0.2}$} & \cellcolor{blue!10}\textcolor{black}{$39.3_{0.2}$} \\
\midrule
\multicolumn{2}{c}{$\Delta_{\mathrm{model}^\dagger - \mathrm{model}^\ast}$} & \textcolor{ForestGreen}{$\uparrow1.5$}  & \textcolor{red}{$\downarrow3.5$}  & \textcolor{blue}{$0.0$}  & \textcolor{ForestGreen}{$\uparrow6.$}  & \textcolor{ForestGreen}{$\uparrow6.1$}  & \textcolor{ForestGreen}{$\uparrow4.$}\\
\bottomrule
\end{tabular}%
}
\end{center}\caption{Detailed results per label for depression}
\label{tab:results_depression}
\end{table}

\begin{table}[!ht]
\begin{center}
\resizebox{\columnwidth}{!}{%
\begin{tabular}{lccccccc}
\toprule
\multirow{2}{*}{\textbf{Model}} & \multirow{2}{*}{\textbf{Type}} & 
\multicolumn{3}{c}{\textbf{Anxiety}} & 
\multicolumn{3}{c}{\textbf{Control}}\\
\cmidrule(lr){3-5}
\cmidrule(lr){6-8}
 &  & \textbf{Precision} & \textbf{Recall} & \textbf{F1} & \textbf{Precision} & \textbf{Recall} & \textbf{F1} \\
\midrule
Mental-BERT$_{\mathrm{base}}^\ast$ & SFT & \cellcolor{blue!45}\textcolor{black}{\underline{$\boldsymbol{54.7_{1.9}}$}} & \cellcolor{blue!30}\textcolor{black}{$64.9_{7.7}$} & \cellcolor{blue!35}\textcolor{black}{$59.2_{3.7}$} & \cellcolor{blue!50}\textcolor{black}{$88.0_{2.1}$} & \cellcolor{blue!45}\textcolor{black}{$82.6_{2.1}$} & \cellcolor{blue!50}\textcolor{black}{\underline{$\boldsymbol{85.2_{0.8}}$}} \\
Mental-BERT$_{\mathrm{large}}$ & SFT & \cellcolor{blue!40}\textcolor{black}{$51.8_{2.0}$} & \cellcolor{blue!25}\textcolor{black}{$58.0_{9.1}$} & \cellcolor{blue!30}\textcolor{black}{$54.3_{4.5}$} & \cellcolor{blue!50}\textcolor{black}{$86.0_{2.2}$} & \cellcolor{blue!45}\textcolor{black}{$82.5_{2.9}$} & \cellcolor{blue!50}\textcolor{black}{$84.1_{0.8}$} \\
Mental-RoBERTa$_{\mathrm{large}}$ & SFT & \cellcolor{blue!40}\textcolor{black}{$51.0_{1.7}$} & \cellcolor{blue!25}\textcolor{black}{$59.8_{8.9}$} & \cellcolor{blue!30}\textcolor{black}{$54.8_{4.3}$} & \cellcolor{blue!50}\textcolor{black}{$86.4_{2.2}$} & \cellcolor{blue!40}\textcolor{black}{$81.5_{2.7}$} & \cellcolor{blue!50}\textcolor{black}{$83.8_{0.7}$} \\
Mental-XLNet$_{\mathrm{base}}$ & SFT & \cellcolor{blue!40}\textcolor{black}{$52.5_{1.6}$} & \cellcolor{blue!25}\textcolor{black}{$59.8_{8.6}$} & \cellcolor{blue!35}\textcolor{black}{$55.6_{3.9}$} & \cellcolor{blue!50}\textcolor{black}{$86.5_{2.1}$} & \cellcolor{blue!45}\textcolor{black}{$82.5_{2.9}$} & \cellcolor{blue!50}\textcolor{black}{$84.4_{0.8}$} \\
Mental-LongFormer$_{\mathrm{base}}$ & SFT & \cellcolor{blue!45}\textcolor{black}{$53.3_{1.8}$} & \cellcolor{blue!30}\textcolor{black}{$61.7_{8.8}$} & \cellcolor{blue!35}\textcolor{black}{$56.9_{4.1}$} & \cellcolor{blue!50}\textcolor{black}{$87.1_{2.4}$} & \cellcolor{blue!45}\textcolor{black}{\underline{$\boldsymbol{82.6_{2.4}}$}} & \cellcolor{blue!50}\textcolor{black}{$84.7_{0.7}$} \\
\midrule
\multirow{2}{*}{LLama-2-chat-7B} & zero-shot & \cellcolor{blue!5}\textcolor{black}{$24.5_{0.1}$} & \cellcolor{blue!50}\textcolor{black}{$98.8_{0.1}$} & \cellcolor{blue!10}\textcolor{black}{$39.2_{0.1}$} & \cellcolor{blue!50}\textcolor{black}{$79.7_{4.6}$} & \cellcolor{blue!5}\textcolor{black}{$1.7_{0.6}$} & \cellcolor{blue!5}\textcolor{black}{$3.2_{1.2}$} \\
& few-shot & \cellcolor{blue!5}\textcolor{black}{$25_{0.5}$} & \cellcolor{blue!20}\textcolor{black}{$65.2_{1.2}$} & \cellcolor{blue!10}\textcolor{black}{$36.2_{0.7}$} & \cellcolor{blue!40}\textcolor{black}{$76.7_{0.9}$} & \cellcolor{blue!15}\textcolor{black}{$36.9_{1.2}$} & \cellcolor{blue!20}\textcolor{black}{$49.8_{1.2}$} \\
\cmidrule(lr){1-8}
\multirow{2}{*}{LLama-2-chat-13B} & zero-shot & \cellcolor{blue!10}\textcolor{black}{$26.4_{0.1}$} & \cellcolor{blue!45}\textcolor{black}{$81.6_{0.7}$} & \cellcolor{blue!10}\textcolor{black}{$39.9_{0.2}$} & \cellcolor{blue!45}\textcolor{black}{$81.6_{0.4}$} & \cellcolor{blue!10}\textcolor{black}{$26.3_{0.8}$} & \cellcolor{blue!10}\textcolor{black}{$39.8_{0.9}$} \\
& few-shot & \cellcolor{blue!5}\textcolor{black}{$23.9_{1}$} & \cellcolor{blue!10}\textcolor{black}{$38.7_{1.3}$} & \cellcolor{blue!5}\textcolor{black}{$29.5_{1.2}$} & \cellcolor{blue!40}\textcolor{black}{$75.2_{0.7}$} & \cellcolor{blue!30}\textcolor{black}{$60.1_{0.9}$} & \cellcolor{blue!35}\textcolor{black}{$66.8_{0.8}$} \\
\cmidrule(lr){1-8}
\multirow{2}{*}{GPT-3.5-turbo} & zero-shot & \cellcolor{blue!10}\textcolor{black}{$28.6_{0}$} & \cellcolor{blue!50}\textcolor{black}{$94.4_{0.1}$} & \cellcolor{blue!15}\textcolor{black}{$43.9_{0.1}$} & \cellcolor{blue!50}\textcolor{black}{$93.0_{0.1}$} & \cellcolor{blue!10}\textcolor{black}{$24.0_{0.1}$} & \cellcolor{blue!10}\textcolor{black}{$38.1_{0.2}$} \\
& few-shot & \cellcolor{blue!10}\textcolor{black}{$27.8_{0}$} & \cellcolor{blue!50}\textcolor{black}{$95.5_{0.1}$} & \cellcolor{blue!15}\textcolor{black}{$43.1_{0.1}$} & \cellcolor{blue!50}\textcolor{black}{$93.2_{0.1}$} & \cellcolor{blue!5}\textcolor{black}{$20.0_{0.2}$} & \cellcolor{blue!5}\textcolor{black}{$32.9_{0.3}$} \\
\cmidrule(lr){1-8}
\multirow{2}{*}{GPT-4} & zero-shot$^\dagger$ & \cellcolor{blue!40}\textcolor{black}{$52.5_{0.1}$} & \cellcolor{blue!45}\textcolor{black}{$83.0_{0.0}$} & \cellcolor{blue!50}\textcolor{black}{\underline{$\boldsymbol{64.4_{0.1}}$}} & \cellcolor{blue!50}\textcolor{black}{\underline{$\boldsymbol{93.3_{0.0}}$}} & \cellcolor{blue!45}\textcolor{black}{$75.8_{0.1}$} & \cellcolor{blue!50}\textcolor{black}{$83.6_{0.1}$} \\
& few-shot & \cellcolor{blue!30}\textcolor{black}{$42.8_{0.2}$} & \cellcolor{blue!50}\textcolor{black}{$90.6_{0.2}$} & \cellcolor{blue!45}\textcolor{black}{$58.1_{0.2}$} & \cellcolor{blue!50}\textcolor{black}{$95.3_{0.1}$} & \cellcolor{blue!35}\textcolor{black}{$60.9_{0.2}$} & \cellcolor{blue!45}\textcolor{black}{$74.3_{0.2}$} \\
\midrule
\multicolumn{2}{c}{$\Delta_{\mathrm{model}^\dagger - \mathrm{model}^\ast}$} & \textcolor{red}{$\downarrow2.2$}  & \textcolor{ForestGreen}{$\uparrow18.1$}  & \textcolor{ForestGreen}{$\uparrow5.2$}  & \textcolor{ForestGreen}{$\uparrow5.3$}  & \textcolor{red}{$\downarrow6.8$}  & \textcolor{red}{$\downarrow1.6$}\\
\bottomrule
\end{tabular}%
}
\end{center}\caption{Detailed results per label for anxiety.}
\label{tab:results_anxiety}
\end{table}

\section{Experimental Setup}

\subsection{Models}
We benchmark \dataset\ using both discriminative and generative language models. The former include Mental-BERT$_{\mathrm{base}}$, Mental-BERT$_{\mathrm{large}}$, Mental-RoBERTa$_{\mathrm{large}}$, Mental-XLNet$_{\mathrm{base}}$, and Mental-LongFormer$_{\mathrm{base}}$ \cite{ji-etal-2022-mentalbert, ji2023domain}. These models have undergone continued pre-training \cite{gururangan-etal-2020-dont} on extensive mental health-related data collected from Reddit, demonstrating state-of-the-art performance on various mental health corpora. Amongst generative language models, we report results on Llama-2 chat models (7B and 13B) \cite{touvron2023llama}, GPT3.5-turbo \cite{ouyang2022-instuctGPT}, and GPT-4 \cite{achiam2023gpt}. We report results for both the zero and few-shot settings. 

\subsection{Evaluation Criteria} 
We benchmark \dataset\ on two tasks, providing alternative but complementary viewpoints for our analysis. They are as follows: 
\begin{itemize}[leftmargin=*]
    \item \textbf{Multi-label classification} where a post can simultaneously be classified as depression, anxiety, both, or None.
    \item \textbf{Binary classification} for depression vs non-depression and anxiety vs non-anxiety.
\end{itemize}

We report the standard weighted precision, recall, and F1 scores for both multi-label and binary classification tasks. Additionally, we provide macro-F1 scores for both tasks. In the case of multi-label classification, we also report the Hamming loss \cite{tsoumakas2007multi}, which measures the fraction of labels for which the predicted and actual labels disagree.It is calculated as the average number of incorrect labels divided by the total number of labels.
\begin{table*}[!ht]
\begin{center}
\resizebox{\textwidth}{!}{%
\begin{tabular}{lcccccccc}
\toprule
\textbf{Method} & \textbf{Prompt} & \textbf{Hamming Loss} & \textbf{Precision} & \textbf{Recall} & \underline{\textbf{F1}} & \textbf{macro-F1} & \textbf{depression F1} & \textbf{anxiety F1} \\ 
\midrule
Mental-BERT$_{\mathrm{large}}$ & SFT & \cellcolor{green!25}\textcolor{black}{$28.3_{7.7}$} & \cellcolor{green!35}\textcolor{black}{$59.4_{8.1}$} & \cellcolor{green!45}\textcolor{black}{$85.2_{15.7}$} & \cellcolor{green!40}\textcolor{black}{$70.0_{8.8}$} & \cellcolor{green!35}\textcolor{black}{$65.2_{13.4}$} & \cellcolor{green!30}\textcolor{black}{$53.3_{2.6}$} & \cellcolor{green!10}\textcolor{black}{$16.4_{7.9}$}  \\
Mental-RoBERTa$_{\mathrm{large}}$ & SFT & \cellcolor{green!25}\textcolor{black}{$28.3_{3.6}$} & \cellcolor{green!35}\textcolor{black}{$59.4_{4.0}$} & \cellcolor{green!45}\textcolor{black}{$84.5_{14.6}$} & \cellcolor{green!40}\textcolor{black}{$69.7_{7.5}$} & \cellcolor{green!35}\textcolor{black}{$64.6_{10.1}$} & \cellcolor{green!30}\textcolor{black}{$53.5_{3.0}$} & \cellcolor{green!10}\textcolor{black}{$16.0_{5.4}$} \\
Mental-XLNet$_{\mathrm{base}}$ & SFT & \cellcolor{green!30}\textcolor{black}{\underline{$\boldsymbol{28.1_{4.5}}$}} & \cellcolor{green!40}\textcolor{black}{\underline{$\boldsymbol{59.6_{4.5}}$}} & \cellcolor{green!45}\textcolor{black}{$84.7_{15.0}$} & \cellcolor{green!40}\textcolor{black}{$69.9_{7.7}$} & \cellcolor{green!35}\textcolor{black}{$64.8_{12.0}$} & \cellcolor{green!35}\textcolor{black}{\underline{$\boldsymbol{53.7_{1.6}}$}} & \cellcolor{green!10}\textcolor{black}{$16.0_{7.1}$} \\
Mental-LongFormer$_{\mathrm{base}}$ & SFT & \cellcolor{green!25}\textcolor{black}{$28.4_{6.5}$} & \cellcolor{green!35}\textcolor{black}{$59.3_{7.2}$} & \cellcolor{green!45}\textcolor{black}{$84.7_{18.7}$} & \cellcolor{green!40}\textcolor{black}{$69.8_{10.3}$} & \cellcolor{green!35}\textcolor{black}{$64.8_{15.4}$} & \cellcolor{green!30}\textcolor{black}{$53.4_{2.8}$} & \cellcolor{green!10}\textcolor{black}{$16.1_{9.0}$}\\
\midrule
\multirow{2}{*}{LLama-2-chat-7B} & zero-shot & \cellcolor{green!5}\textcolor{black}{$52.5_{0.1}$} & \cellcolor{green!10}\textcolor{black}{$51.3_{0.2}$} & \cellcolor{green!50}\textcolor{black}{$96.9_{0.4}$} & \cellcolor{green!30}\textcolor{black}{$65_{0.1}$} & \cellcolor{green!20}\textcolor{black}{$58_{0.1}$} & \cellcolor{green!30}\textcolor{black}{$52.7_{0}$} & \cellcolor{green!5}\textcolor{black}{$12.3_{0.1}$} \\
& few-shot & \cellcolor{green!5}\textcolor{black}{$52.9_{0.7}$} & \cellcolor{green!5}\textcolor{black}{$48.6_{0.5}$} & \cellcolor{green!45}\textcolor{black}{$81.4_{1.1}$} & \cellcolor{green!20}\textcolor{black}{$59.1_{0.7}$} & \cellcolor{green!10}\textcolor{black}{$53.3_{0.6}$} & \cellcolor{green!10}\textcolor{black}{$47.4_{0.6}$} & \cellcolor{green!5}\textcolor{black}{$11.7_{0.2}$} \\
\cmidrule(lr){1-9}
\multirow{2}{*}{LLama-2-chat-13B} & zero-shot & \cellcolor{green!10}\textcolor{black}{$50.4_{0.3}$} & \cellcolor{green!10}\textcolor{black}{$52.2_{0.1}$} & \cellcolor{green!50}\textcolor{black}{$95.1_{0.5}$} & \cellcolor{green!30}\textcolor{black}{$65.3_{0.2}$} & \cellcolor{green!20}\textcolor{black}{$58.5_{0.3}$} & \cellcolor{green!30}\textcolor{black}{$52.8_{0.2}$} & \cellcolor{green!5}\textcolor{black}{$12.5_{0.1}$} \\
& few-shot & \cellcolor{green!5}\textcolor{black}{$44.3_{1.1}$} & \cellcolor{green!5}\textcolor{black}{$47.5_{1.8}$} & \cellcolor{green!5}\textcolor{black}{$28.1_{1}$} & \cellcolor{green!5}\textcolor{black}{$34.2_{1}$} & \cellcolor{green!5}\textcolor{black}{$32.4_{1.3}$} & \cellcolor{green!5}\textcolor{black}{$25.6_{0.4}$} & \cellcolor{green!5}\textcolor{black}{$8.6_{0.7}$} \\
\cmidrule(lr){1-9}
\multirow{2}{*}{GPT-3.5-turbo} & zero-shot & \cellcolor{green!10}\textcolor{black}{$46.1_{10.2}$} & \cellcolor{green!15}\textcolor{black}{$51.4_{5.3}$} & \cellcolor{green!50}\textcolor{black}{$92.1_{9.4}$} & \cellcolor{green!35}\textcolor{black}{$67.5_{4.5}$} & \cellcolor{green!25}\textcolor{black}{$60.5_{4.5}$} & \cellcolor{green!30}\textcolor{black}{$52.3_{3.1}$} & \cellcolor{green!10}\textcolor{black}{$14.0_{1.4}$} \\
& few-shot & \cellcolor{green!25}\textcolor{black}{$30.3_{11.7}$} & \cellcolor{green!35}\textcolor{black}{$57.8_{8.0}$} & \cellcolor{green!50}\textcolor{black}{\underline{$\boldsymbol{98.4_{0.7}}$}} & \cellcolor{green!40}\textcolor{black}{\underline{$\boldsymbol{71.0_{3.9}}$}} & \cellcolor{green!40}\textcolor{black}{\underline{$\boldsymbol{67.4_{5.6}}$}} & \cellcolor{green!35}\textcolor{black}{$53.0_{0.9}$} & \cellcolor{green!10}\textcolor{black}{\underline{$\boldsymbol{17.9_{3.1}}$}} \\
\cmidrule(lr){1-9}
\multirow{2}{*}{GPT-4} & zero-shot & \cellcolor{green!30}\textcolor{black}{$34.0_{5.2}$} & \cellcolor{green!35}\textcolor{black}{$55.6_{2.2}$} & \cellcolor{green!50}\textcolor{black}{$97.4_{0.3}$} & \cellcolor{green!40}\textcolor{black}{$70.8_{2.2}$} & \cellcolor{green!40}\textcolor{black}{$66.7_{3.9}$} & \cellcolor{green!35}\textcolor{black}{$52.9_{0.5}$} & \cellcolor{green!10}\textcolor{black}{$17.5_{2.5}$} \\
& few-shot & \cellcolor{green!30}\textcolor{black}{$35.6_{15.1}$} & \cellcolor{green!35}\textcolor{black}{$54.8_{5.7}$} & \cellcolor{green!50}\textcolor{black}{$97.4_{3.5}$} & \cellcolor{green!40}\textcolor{black}{$70.1_{5.1}$} & \cellcolor{green!35}\textcolor{black}{$65.7_{8.6}$} & \cellcolor{green!35}\textcolor{black}{$52.6_{0.3}$} & \cellcolor{green!10}\textcolor{black}{$17.0_{5.4}$} \\
\midrule
\multicolumn{2}{c}{$\Delta_{\mathrm{model}^\dagger - \mathrm{model}^\ast}$} & \textcolor{red}{$\uparrow2.2$} & \textcolor{red}{$\downarrow1.8$} & \textcolor{ForestGreen}{$\uparrow1.0$} & \textcolor{ForestGreen}{$\uparrow0.02$} & \textcolor{ForestGreen}{$\uparrow0.7$} & \textcolor{red}{$\downarrow0.7$} & \textcolor{ForestGreen}{$\uparrow0.4$} \\
\bottomrule
\end{tabular}%
}
\end{center}\caption{Results for depression and anxiety multi-label classification}
\label{tab:results_comorbidity}
\end{table*}

\section{Results and Discussion}
\label{sec:results_and_discussion}


\subsection{Binary Classification.}

The overall binary classification results for depression and anxiety are presented in Table \ref{tab:results_binary}. We observe that discriminative models exhibit consistent performance across both tasks, outperforming much larger models such as LLama-2 and GPT3.5-turbo. Apart from GPT-4, all other generative models show considerable variance in performance in both zero- and few-shot settings. Furthermore, there is a noticeable difference in how models handle depression versus anxiety. 

\paragraph{Depression results} --  The fine-grained results for depression classification presented in Table \ref{tab:results_depression}, demonstrate a differential performance across labels. Models consistently exhibit higher recall and F1 scores for the depression label compared to the control group, indicating a stronger ability to identify true cases of depression while accepting lower precision. This trend signifies a model's bias towards minimizing false negatives, thus reducing the risk of missing actual depression cases. The comparison between discriminative models and GPT-4 shows only a slight difference. Both GPT-4 and Mental-XLNet achieve similar F1 scores; however, GPT-4 exhibits superior performance in classifying the control groups. This suggests that while the models are effective in identifying depression, there exists a precision trade-off that prioritizes avoiding missed diagnoses.

\paragraph{Anxiety results} -- Table \ref{tab:results_anxiety} shows the fine-grained results for anxiety classification. GPT-4 significantly outperforms all other discriminative and generative models in both zero-shot and few-shot settings, achieving the highest F1-scores for both anxiety ($64.4\%$) and control ($83.6\%$). Generally, all models demonstrate better performance for the control label across precision, recall, and F1-score compared to the anxiety label, indicating a consistent pattern of effectively identifying non-anxiety cases. Furthermore, there is a notable discrepancy between the precision and recall for the anxiety label, with precision significantly lower than recall. This discrepancy suggests a high rate of false positives, indicative of Type I errors. In practical diagnostic settings, this could lead to over-diagnosis. A possible cause for this issue is the imbalance in \dataset\, where anxiety cases are underrepresented compared to control cases, potentially causing the models to predict anxiety more frequently to capture most true anxiety instances.

\subsection{Multi-Label Classification.}
Table \ref{tab:results_comorbidity} presents the results for multi-label classification. We observe a consistent trend of high recall and moderate precision. This indicates that while the models excel at identifying relevant labels and minimizing false negatives, they incurr high false positives. Notably, the models demonstrate superior performance in predicting depression compared to anxiety, as evidenced by consistently higher F1 scores for depression. This disparity could stem from the more pronounced or readily learnable language features associated with depression in the dataset. GPT-3.5 Turbo with few-shot prompting stands out as the most effective, achieving an overall balanced F1 score of $71\%$, with noteworthy scores of $53\%$ for depression and $17.9\%$ for anxiety. 
On the other hand, PLMs that are characterized by lower Hamming Loss, adopt a more conservative prediction approach, likely reducing false positives but potentially missing some true positives. In contrast, GPT-4, with its higher F1 scores and greater Hamming Loss, adopts a more liberal approach, enhancing its ability to capture true positives at the expense of increasing mis-classification rates.

\section{Error Analysis}
\label{sec:error_analysis}

\subsection{Few shot vs Zero Shot}
From our qualitative analysis, we observe that zero-shot prompting outperforms few-shot learning. Out of $334$ samples where the predictions of the two approaches differed, the zero-shot prompt correctly classified $241$ samples that the few-shot prompt misclassified. Only $4/241$ samples were actually labeled as depressed in the ground truth, while the remaining $237$ were non-depressed instances. Notably, all $4$ of these depressed examples contained self-diagnosis statements, which the few-shot approach failed to identify correctly (see Table \ref{tab:sA}). For anxiety classification, out of the $468$ samples where the predictions differed, the zero-shot approach correctly classified $363$ non-anxiety samples that the few-shot approach misclassified. We observe that slight anxious behavior mentioned in the posts is aggressively classified as an anxiety disorder by the few-shot approach, despite the ground truth indicating non-anxiety

Upon qualitative analysis of these misclassified samples, we hypothesize that the few-shot prompts may be inducing noise and bias by overgeneralizing from the limited examples provided in context. Specifically, many non-depressed posts contain mentions of depressive symptoms, which the few-shot model appears to heavily weight, leading to false positive predictions. This can also be attributed to the fact that the in-context learning examples are silver labeled, the retriever is not potent enough to understand the minutiae in the semantics, or LLM’s ability to handle long context.

\subsection{Error-Analysis -- The Battle of the GPTs}
Both GPT-3.5-turbo and GPT-4 show a limitation in assessing the temporal aspect of depression. Among the $761$ samples misclassified by both models, only $42$ samples with a true label of "depressed" were incorrectly labeled as "not depressed." The models struggled to differentiate between past and present states of depression (Refer Table \ref{tab:sB1}). 
Focusing on zero-shot depression, a total of $334$ samples differed for GPT-3.5-turbo vs GPT-4. Out of these, GPT-4 demonstrated superior performance, with $270$ samples correctly classified, while GPT-3.5 provided incorrect classifications. Among the $270$ samples where GPT-4 was accurate, 251 instances were correctly identified as non-depressed, contradicting GPT-3.5's classification as depressed. This finding highlights GPT-4's enhanced contextual understanding and ability to differentiate between genuine depressive symptoms and non-depressive states, thus avoiding the over-classification of non-depressive instances as depressed.
\section{Conclusion}
In this study, we propose \dataset\, a novel dataset aimed at bridging critical gaps in the comorbid diagnosis of depression and anxiety from social media posts. By embracing a multi-label classification approach, \dataset\ promotes research in early detection and understanding of comorbid depression and anxiety conditions, which are often co-occurring yet underserved by existing datasets. Our benchmarking experiments with state-of-the-art language models, including both discriminative and generative methods, highlight the superior performance of GPT-4 in both binary and multi-label classification tasks, reflecting its advanced capabilities in symptom identification and context understanding. Conversely, LLMs such as GPT-3.5 and LLaMa underperform relative to domain-specific PLMs like Mental-XLM, underscoring the inherent challenges and limitations of these models in handling the complexity of mental health diagnosis from text. However, despite these advancements, all models demonstrated a tendency to misclassify complex cases, indicating a need for improved methods that can differentiate the subtleties of mental health language and symptom expressions, especially in settings where disorders co-occur. In the future, we aim to explore hybrid models that combine the strengths of both discriminative and generative approaches to improve the accuracy and reliability of mental health assessments based on social media data, ultimately contributing to the development of more effective tools for early diagnosis and intervention, thus improving outcomes for individuals with mental health needs.
\section{Limitations}

This study has several limitations that future research could address. Firstly, our data collection from Reddit was constrained by time limitations and a finite search space, leading to potential blind spots in identifying users' self-disclosure. There may be rare expressions of self-disclosure or posts from non-mental health related subreddits were we missed curating \dataset . Additionally, our dataset only includes text, omitting other potentially informative modalities. For instance, the timing of posts could indicate insomnia or social relationship issues, which are early signs of depression or anxiety. We also limit our study to two disorders - depression and anxiety, and include posts in English only. Moreover, as discussed in section \ref{sec:results_and_discussion}, the performance of all baseline models in classification tasks is far from ideal. Both generative and discriminative models, including GPT-4, consistently exhibit low precision, which shows that they are prone to false diagnoses. Therefore, while these models could serve as preliminary tools for individuals unaware of their mental conditions or those unable to access mental health services, their predictions must be rigorously reviewed by professionals before confirming a diagnosis.
\section{Ethical Considerations}
Social media data can be highly sensitive, especially when pertaining to mental health. Thus, it is imperative to prioritize privacy and recognize the potential risks posed to individuals represented within the data \cite{hovy-spruit-2016-social, suster-etal-2017-short, benton-etal-2017-ethical}. Given this consideration, we assert that the risks tied to the data used in this study are minimal. Our assessment is corroborated by prior studies which have introduced similar datasets \cite{coppersmith-etal-2015-clpsych, milne-etal-2016-clpsych,losada2016test}

The \dataset\ dataset solely consists of publicly accessible Reddit posts. We diligently remove any information that could disclose an author's identity or demographics. We provided annotators only with anonymized posts and ensured their commitment to neither deanonymize nor contact the authors. Moreover, we extensively adhere to the ethical and privacy guidelines set forth by \cite{benton-etal-2017-ethical}. We do not collect any identifiable information, and we securely store all data on protected servers, accessible solely through written agreements with the creators. The institutional review board (IRB) at our institution has classified our experiments using these datasets as exempt from additional review.

\bibliography{anthology, custom}
\clearpage
\newpage
\appendix
\section{Appendix}
\label{sec:appendix}
\subsection{Curating \Silverdataset}
\label{sec:silver_data}

Since \dataset\ serves exclusively as a test benchmark, we compiled \Silverdataset\ as a complementary corpus suitable for few-shot learning or supervised fine-tuning. Given the complexity and length of the posts, which demand considerable human and financial resources, we utilized LLMs for annotation. Recent advancements in GPT-based solutions have shown them to be effective substitutes for human labeling, aligning well with human judgments in both clinical and non-clinical tasks \cite{wang2021-GPT3-silver-label, li-etal-2023-GPT-silver-labels, du-etal-2023-GPT-silver-labels, he2024annollm-silver-labelling}. Furthermore, \citet{li-etal-2023-GPT-silver-labels} demonstrated that generating explanations alongside target labels enhances the quality of silver labels, achieving a higher correlation with crowd-sourced annotations. Leveraging this approach, we employed GPT-3.5-turbo \cite{ouyang2022-instuctGPT} to generate silver labels for the remaining $22,124$ posts in a zero-shot setting. Using an open-ended prompt, as detailed in Table \ref{tab:silver_label_prompt}, we directed the LLM to identify and rationalize cues related to any mental health disorder that may be attributed with the post. After analyzing GPT-3.5-turbo outputs, we retained $7,667$ posts identified with depression and/or anxiety and discarded the rest, thus forming \Silverdataset.

\begin{table*}[!ht]
\begin{center}
\resizebox{\textwidth}{!}{%
\begin{tabular}{p{7.2in}cccccc}
\toprule
\textbf{Post} & \textbf{NRC (anger)} & \textbf{NRC (disgust)} & \textbf{NRC (fear)} & \textbf{NRC (sadness)} & \textbf{NRC (total)} & \textbf{Empath (total)} \\
\midrule
\textit{\textcolor{black}{hi everyone! im a little late in the tour game and i didnt realize i wanted to see ari until this past week. now that shes announced shes adding more us dates, however, im hoping that she comes somewhere close to me so i get the chance to see her.} the only ticket prices ive seen are resale, of course, for current shows. i was wondering if anyone remembers how much certain tickets cost (cheapest tickets, floor seats, pit, soundcheck, mg); these are all ball park of course since each venue is different. if you went to the cleveland, detroit, or pittsburgh show and remember prices for each section thats a bonus. thank you!} & $0.000$ & $0.000$ & $0.000$ & $0.125$ & $0.125$ & $0.111$ \\
\cmidrule(lr){1-7}
\textit{I've got a bunch of academies for my knights in a kingdom with a currently unknown population. I'm trying to figure out how many people I should have based on lifespan and death rates but I also need to know how believable it would be to have student base of, say, 1000. Before college I never went to school so I just don't know what is abnormal or what would make people stop and question the population, but I feel it's an important detail. For some background, the average human lifespan is 1000 years. Roughly 40\% of the population are knights. Of that 40\%, and I never learned even basic math so I'm just throwing out numbers here, let's say 5\% are students. \textcolor{red}{Would it be reasonable to have a smaller student base then? I would say with a low birth rate it would be a better option.}} & $0.574$ & $1.094$ & $1.415$ & $0.915$ & $3.998$ & $0.333$ \\
\cmidrule(lr){1-7}
\textit{\textcolor{red}{I hate the thought that I might be becoming antisocial, but I am enjoying my alone time too much. I've lived alone for most of the past few years at university, and now I'm back home temporarily whilst trying to get a decent job. The weekends take it out of me because I've got to be social and interact with my family otherwise I feel terrible inside because they like my company and it's not fair of me to constantly be hauled up in my room. Then Mondays come and I'm relieved when everybody goes to work, because I have time to myself.} \textcolor{red}{Problem is I get extremely irritable when I don't get that time.} or I plan time to myself or it doesn't work out. For example, my dad's not going to work today most likely, because his van won't start. I love my dad and I love spending time with him when I've prepared for it, but \textcolor{red}{when something sudden like this happens I feel cheated out of my alone-time.} Even when I'm not hanging out with him and I'm upstairs by myself, \textcolor{red}{I still feel very anxious and I feel bad for not interacting.} Anybody else have this problem? How'd you force yourself into those interactions, or do you just give into your need to be alone? My family often worry I'm depressed (or *more* depressed) and I want to prove them wrong but \textcolor{red}{I also want to be alone.}} & $3.987$ & $2.179$ & $3.925$ & $4.903$ & $14.994$ & $1.778$ \\
\bottomrule
\end{tabular}%
}
\end{center}
\caption{NRC emotion and Empath scores for sample posts with highlighted text reflecting the NRC values.}
\label{tab:nrc_empath}
\end{table*}

\begin{table}[ht!]
\begin{center}
\small 
\resizebox{0.95\columnwidth}{!}{
\begin{tabular}{lcccc}
\toprule
\multirow{2}{*}{\textbf{Source}} & \multicolumn{4}{c} {\textbf{Evaluation Metrics}} \\
\cmidrule(lr){2-5}
& \textbf{Acc} & \textbf{F1} & \textbf{ROC-AUC} & \textbf{MCC}\\
\midrule
\small
DATD & $0.956$ & $0.933$ & $0.904$ & $0.875$ \\
Dep Reddit & $0.939$ & $0.938$ & $0.937$ & $0.878$ \\
Dreaddit & $0.925$ & $0.924$ & $0.923$ & $0.857$ \\
SDCNL & $1.000$ & $1.000$ & $1.000$ & $0.990$ \\
\bottomrule
\end{tabular}%
}
\end{center}
\caption{Adversarial validation using \dataset\  as the target dataset against contemporary datasets as the source. Higher values of the metrics indicate a greater degree of discriminatory power of \dataset\ in comparison to the source datasets.}
\label{tab:adv}
\end{table}

\subsection{Inter-Class Similarity -- Case study of MMD}
\label{sec:mmd}
We report the mean MMD over 1000 runs with a random label-wise pair chosen at each run without replacement. In \dataset, we observed that the MMD between the control group and the groups representing anxiety, depression, and comorbidity is notably higher (0.47) than the MMD among the latter groups themselves (0.42). This observation suggests that distinguishing between anxiety, depression, and comorbidity poses a relatively greater challenge compared to distinguishing the control group from the other classes. Furthermore, when we compare the MMD of \dataset\  to other datasets, we find that our dataset exhibits a significantly lower MMD. This observation corroborates our hypothesis that \dataset\ presents a higher difficulty level in classification tasks due to the increased proximity between its class distributions.

\subsection{Hyperparameter Tuning}
Ww fine-tune the BERT-based models on \Silverdataset. We vary the learning rate within the $\{2-5, 2e-6\}$ range and experiment with batch sizes of $\{4, 8, 16, 32\}$. We train each model for $30$ epochs with early stopping applied after $10$ epochs of no improvement in validation F1 score. We use the Adam optimizer with decoupled weight decay regularization of $10^{-2}$ \citep{loshchilov2017decoupled}. For generating the outputs from the LLMs, we use top-p decoding with $p=0.95$ \cite{Holtzman2020The}. For each hyperparameter configuration, we run $3$ seeds and report the averaged results. We apply this hyperparameter setup across depression, anxiety, and comorbidity classification tasks.

We use a single consistent prompt for zero-shot prompting of all the generative language models in our setup, provided in Appendix \ref{sec:prompts}. For few-shot prompting, we utilize the silver-labeled data as the retrieval pool. For each example, we retrieve the two most semantically similar posts from the silver-labeled data, irrespective of its label, to serve as in-context learning examples. We employ the all-mpnet-base-v2 model from sentence-transformers \cite{reimers-gurevych-2019-sentence} for computing semantic similarity between the test example and the silver-labeled data. The few-shot prompt is also included in Section \ref{sec:prompts}.

\subsection{Zero and Few-Shot Prompts}
\label{sec:prompts}
Table \ref{tab:prompts} illustrates the zero and few shot prompts used for binary (depression vs non-depression and anxiety vs non-anxiety) and multi-label classification on \dataset\ using LLMs.

\begin{table*}[!ht]
\begin{center}
\resizebox{\textwidth}{!}{%
\begin{tabular}{c|>{\centering\arraybackslash}m{2.5in}|>{\centering\arraybackslash}m{3.5in}}
\toprule
\textbf{Task} & \textbf{Zero-Shot Prompt} & \textbf{Few-Shot Prompt} \\
\midrule
Depression & \parbox[t]{2.5in}{\texttt{Assess whether the content of the post delimited by triple backticks aligns with the criteria for clinical depression as defined in DSM-5. Format your response as a JSON object {'depression':''} with values strictly either 'yes' or 'no'. \\ Do not return anything else apart from the json object.\\ Post: \\ ```}} & \parbox[t]{3.5in}{\texttt{Below are posts and their respective assessments based on the criteria for clinical depression as defined in the DSM-5.\\ Format your response as a JSON object {'depression':''} with values either 'yes' or 'no'. \\
Post: <post\_1> \\
Assesement: <assessment\_1> \\
Post: <post\_2> \\
Assesement: <assessment\_2> \\
Based on the above, assess the content of the following post:\\
Post: <post\_test>\\
Assessment:}} \\
\midrule
Anxiety & \parbox[t]{2.5in}{\texttt{Assess whether the content of the post delimited by triple backticks aligns with the criteria for clinical anxiety as defined in DSM-5. Format your response as a JSON object {'anxiety':''} with values strictly either 'yes' or 'no'.\\Do not return anything else apart from the json object.\\Post: \\ ```}} & \parbox[t]{3.5in}{\texttt{Below are posts and their respective assessments based on the criteria for clinical anxiety as defined in the DSM-5.\\ Format your response as a JSON object {'anxiety':''} with values either 'yes' or 'no'. \\
Post: <post\_1> \\
Assesement: <assessment\_1> \\
Post: <post\_2> \\
Assesement: <assessment\_2> \\
Based on the above, assess the content of the following post:\\
Post: <post\_test>\\
Assessment:}} \\
\midrule
Comorbidity & \parbox[t]{2.5in}{\texttt{This person wrote this paragraph on social media. As a psychologist, read the post on social media and answer the following questions. \\ 1. Does the poster of this post have anxiety? \\ 2. Does the poster of this post have depression? \\ Format your response as a JSON object {'depression':'', 'anxiety': ''} with values strictly either 'yes' or 'no'. Do not return anything else apart from the json object. \\ Post: \\ ```}} & \parbox[t]{3.5in}{\texttt{Below are posts and their respective assessments based on the criteria for clinical depression and clinical anxiety as defined in the DSM-5. \\ Format your response as a JSON object {'depression': '', 'anxiety': ''} with values either 'yes' or 'no'.\\
Post: <post\_1> \\
Assesement: <assessment\_1> \\
Post: <post\_2> \\
Assesement: <assessment\_2> \\
Based on the above, assess the content of the following post:\\
Post: <post\_test>\\
Assessment:}} \\
\bottomrule
\end{tabular}%
}
\end{center}
\caption{The zero and few-shot prompts used for depression, anxiety, and comorbidity classification.}
\label{tab:prompts}
\end{table*}

\subsection{Other Prompting Techniques}

Self-evaluation scales are commonly used by clinicians to assess the presence and severity of mental disorders based on responses to structured questions. Each scale comprises specific questions, and responses are quantitatively scored to determine the severity of the corresponding mental disorder.

\subsubsection{Depression}

For the assessment of depression, two scales were utilized: PHQ-9 \cite{kroenke2001phq} and MADRS\cite{williams2008development}.

\begin{itemize}[leftmargin=*]
    \item \textbf{PHQ-9:} This scale consists of 9 symptoms, each assessed on a scale ranging from ``not at all" to ``nearly every day," scored from 0 to 4. To streamline responses, we converted these options into binary responses (``yes" or ``no"). The model was prompted to provide a JSON response based on these inputs. Post-response parsing allowed for calculation of the total score, which was then evaluated against a predefined threshold for depression.
    
    \item \textbf{MADRS:} Comprising 10 symptoms, each rated on a scale of 0 to 6 for severity, this scale was adapted in our prompt to request the model to score each symptom from 1 to 6. Post-prompt parsing facilitated computation of the total score, which was subsequently compared against the specified threshold for depression.
\end{itemize}

\subsubsection{Anxiety}

For anxiety assessment, the scales chosen were BAI \cite{beck1993beck} and Hamilton Anxiety \cite{maier1988hamilton} Rating Scale.

\begin{itemize}[leftmargin=*]
    \item \textbf{Hamilton Anxiety Rating Scale:} This scale involves 14 questions, each scored from 0 to 4, evaluating various anxiety symptoms. The model was prompted to assign a score to each symptom, and the cumulative score was calculated to determine anxiety severity.
    
    \item \textbf{BAI (Beck Anxiety Inventory):} Consisting of 21 symptoms, each rated from 0 to 4, this scale was utilized by prompting the model to assign scores to each symptom. Post-prompt parsing of the model's output allowed for computation of the total score, enabling assessment against established anxiety thresholds.
\end{itemize}

\subsubsection{Results}

\textbf{Depression: } Table~\ref{tab:scales_depression} shows the classification scores for depression using MADRS and PHQ-9 (\cite{kroenke2001phq}) scales. The F1 score for depression is generally higher than the control F1 score. For the MADRS prompt, there is an improvement in both Depression F1 and Control F1 when moving from GPT-3.5 to GPT-4. However, for the PHQ-9 scale, the Depression F1 score drops significantly, while the Control F1 is significantly increased. Overall, the GPT-4 results are well-balanced, whereas the GPT-3.5 results are more biased towards classifying a sample as depressed. It is observed that the model tends to hallucinate while scoring, overestimating the presence of symptoms not mentioned in the post. This effect is reduced in GPT-4 compared to GPT-3.5.\\
\textbf{Anxiety: } Table~\ref{tab:scales_anxiety} presents the classification scores for anxiety using the BAI and Hamilton scales. For GPT-3.5, the Hamilton scale yields more balanced results compared to the BAI scale. The GPT-3.5 scores are reasonable, but GPT-4 significantly underperforms on the Hamilton scale.

\subsubsection{Discussion}
The obtained scores fall short of both zero-shot and few-shot performance. This could be due to several reasons, such as the posts not providing sufficient information to answer all questions in a questionnaire accurately. The model's hallucination has a significant impact on the final score, and this effect becomes more pronounced as the number of questions in the questionnaire increases. Different threshold values have been tried, but the issue of insufficient information in the posts cannot be eliminated. Future researchers could use this questionnaire-based prompt to further develop and improve the approach.

\begin{table}[h!]
\begin{center}
\resizebox{\columnwidth}{!}{%
\begin{tabular}{lccccccc}
\toprule
\multirow{2}{*}{\textbf{Model}} & \multirow{2}{*}{\textbf{Type}} & 
\multicolumn{3}{c}{\textbf{Depression}} & 
\multicolumn{3}{c}{\textbf{Control}}\\
\cmidrule(lr){3-5}
\cmidrule(lr){6-8}
 &  & \textbf{Precision} & \textbf{Recall} & \textbf{F1} & \textbf{Precision} & \textbf{Recall} & \textbf{F1} \\
\midrule
\multirow{2}{*}{GPT-3.5-turbo} & MADRS & $57.6$ & $91.1$ & $70.6$ & $62.5$ & $18.2$ & $28.2$ \\
& PHQ-9 & $60.$ & $89.6$ & $72.4$ & $69.$ & $29.2$ & $41.2$ \\
\cmidrule(lr){1-8}
\multirow{2}{*}{GPT-4} & MADRS & $60.3$ & $87.6$ & $71.4$ & $66.0$ & $29.5$ & $40.$ \\
& PHQ-9 & $64.4$ & $46.0$ & $53.$ & $51.1$ & $68.9$ & $58.$ \\
\bottomrule
\end{tabular}%
}
\end{center}\caption{Detailed Results Per Label of Dataset Using PHQ-9 and MADRS Self-Assessment Scales with GPT-3.5 and GPT-4 for Binary Depression Classification}
\label{tab:scales_depression}
\end{table}

\begin{table}[h!]
\begin{center}
\resizebox{\columnwidth}{!}{%
\begin{tabular}{lccccccc}
\toprule
\multirow{2}{*}{\textbf{Model}} & \multirow{2}{*}{\textbf{Type}} & 
\multicolumn{3}{c}{\textbf{Anxiety}} & 
\multicolumn{3}{c}{\textbf{Control}}\\
\cmidrule(lr){3-5}
\cmidrule(lr){6-8}
 &  & \textbf{Precision} & \textbf{Recall} & \textbf{F1} & \textbf{Precision} & \textbf{Recall} & \textbf{F1} \\
\midrule
\multirow{2}{*}{GPT-3.5-turbo} & BAI & $27.2$ & $87.$ & $41.5$ & $85.9$ & $24.2$ & $37.$ \\
& Hamilton & $26.$ & $76.5$ & $39.5$ & $80.9$ & $32.1$ & $46$ \\
\cmidrule(lr){1-8}
\multirow{2}{*}{GPT-4} & BAI & SFT & SFT & SFT & SFT & SFT & SFT\\
& Hamilton & $42.9$ & $6$ & $10.5$ & $76.2$ & $97.4$ & $85.5$ \\
\bottomrule
\end{tabular}%
}
\end{center}\caption{Detailed Results Per Label of Dataset Using BAI and Hamilton Self-Assessment Scales with GPT-3.5 and GPT-4 for Binary Anxiety Classification}
\label{tab:scales_anxiety}
\end{table}

\begin{table*}[!ht]
\centering
\resizebox{\textwidth}{!}{%
\begin{tabular}{p{8.5in}ccc}
\toprule
\textbf{Example} & \textbf{True} & \textbf{Zero-Shot} & \textbf{Few-Shot} \\
\midrule
\textit{MUST SEE LECTURE! Depression is just as biological as Diabetes, by Stanfords Robert Sapolsky. I just watched a lecture by a Stanford Professor, Robert Sapolsky, explaining how the biology and the psychology of depression work together. My \textcolor{red}{BPD is comorbid with a major depressive disorder} and I have been told MANY too many times that I feel miserable because I lack faith, will and overall reasonability. Close relatives and friends Are they friends? regularly compare their everyday stress management to my depression and instruct me on how easy it would be for me to just bounce back. I know I am not the only one to have experienced such ignorance and found reassurance and soothing in listening Professor Sapolsky. Hope it will bring a bit of calm to your days! Would highly recommend it if your BPD is comorbid with depression of if youre trying to understand how your brain is different from a neurotypicals brain. Heres the link Some of my favorite quotestake outs When you want to come to basic meat and potatoes of human medical misery, there is nothing out there like depression Depression is a biochemical disorder with a genetic component and early experience influences where somebody cant appreciate sunsets It can be considered that you are a major risk to yourself when you start getting better after a major episode, not during the episode itself. As you get better you also get the energy to do something catastrophic. This depression screams biology Somewhere around the 4th or 5th major depressive episode, you are statically at a higher risk of relapsing than before. This depression is not Oh, pull yourself together, we all get depressed. This is as real of a biological disorder as is diabetes'} & 1 & 1 & 0 \\
\cmidrule(lr){1-4}
\textit{Ive made myself alone these past few years. I was \textcolor{red}{diagnosed with depression about 1.5 years ago,} and there had been so much going on with my life, struggling with college, my father cheating on my mom, and the collapse of my longdistance relationship. The solution was to change meds, and I finally feel like myself again. I didnt do enough research at the very beginning of my depression, so I took an SSRI, which was the worst thing I could have done looking back. The sideeffects were awful, it was changing who I was, and ultimately, it didnt help me, but at least I didnt feel like killing myself. But, now that Im here, I realize how alone I made myself, given the holidays and the parties that people my age go to. I would have felt like a failure or a loner, but I know that can change. I feel good enough to go out and make friends. If you are isolating yourself, I implore you, from experience, that while you may feel like shit, while you may feel like the lowest thing in the world, there are people that care about you, so much more than you think or feel. Im here now, and Im fighting again, because thats all I can do to make my life better. You can keep fighting, because there are so many people to believe in you. Just something I wanted to get off my chest.} & 1 & 1 & 0\\
\cmidrule(lr){1-4}
\textit{I got out of the Maze. I was \textcolor{red}{diagnosed with depression} at April.For 6 months I felt a void inside me.The last 2 months were horrible.Couldnt move out of the bed.No appetite.No concentration and no memory.Pain everywhere on my body.23 Panic attacks per day.I slept 1 hour each day.Couldnt sleep for more.The anxiety and fear were at 100.Everyday I was watching nightmares.Sucidal thoughts at medium range, but I wanted to die so much from a disease or hit by a car.Started antidepressants at April.The last 3 days are the best days of my life.No depression symptoms and 100 fictional.I do every task and every conversation with so much happiness and joy inside me.I feel I got out of the maze and I am at the sky.I have no specific tip.The most important for my case I think that I never stopped going on my psychologist and psychiatrist.Every week I was there on my sessions,besides how bad I was feeling and how tired I was.I post this to give some hope, you can all do it like me or like anyone else that fought depression.You are no different. P.s. Sorry for my bad English,I am still recovering everyday.} & 1 & 1 & 0\\
\cmidrule(lr){1-4}
\textit{I want to live now and it sucks. \textcolor{red}{I was quite depressed} all through my final year of highschool and most of university. My episodes followed one another with seemingly no end. I was diagnosed with recurring MDD. During that time I was quite suicidal and apathetic and I actually accepted the idea of death. I liked to think death was better than this life and I honestly wouldnt have minded if I died. This was around a year and a half ago. I actually started to recover and with the right meds and therapy I could finally say I was in remission so to speak. The problem is, now Im terrified by the idea of death. Ive been having tiny existential crises one after the other. I cant just accept the void anymore and its honestly making me wish I could go back. Ive even considered one day going off my meds to more easily accept my own death and that of my lived ones. Im still getting used to the feelings and emotions that were gone for such a long time but this bothers me the most and I have no idea how to deal with it.} & 1 & 1 & 0\\
\bottomrule
\end{tabular}%
}
\caption{Instances highlighted in red represent self-diagnosis cases where zero-shot learning correctly classified the instances as depicting depression, while few-shot learning misclassified them, despite the ground truth being depression.}
\label{tab:sA}
\end{table*}

\begin{table*}[!ht]
\centering
\resizebox{\textwidth}{!}{%
\begin{tabular}{p{8.5in}ccc}
\toprule
\textbf{Example} & \textbf{True} & \textbf{GPT3.5} & \textbf{GPT4} \\
\midrule
\textit{Anxiety disorder exacerbating phobias. Background late  \textcolor{red}{20sF diagnosed with anxiety, depression, OCD, and panic disorder at age 11 I have always had a phobia of spiders.} When I was a kid, a nest hatched in my room and there were hundreds of them. It was really traumatizing. I could not sleep in my room for almost a year, out of sheer fear. My husband and I recently moved to a new town, but decided to rent an apartment more on the outskirts and its a bit woodsy. We needed the quiet. However, its only been 3 weeks and we have already seen more critters in our place than we did in 2 years living in a city. One of them was hanging from the ceiling over my head last night and my husband went to kill it and it fell onto me, and I had a complete meltdown. I had a massive panic attack, the worst Ive had in years. I was scratching all over my body and dry heaving and shaking. I thought I was having a heart attack. It was awful. I had nightmares about it in my sleep when I did sleep we had to have the lights on because I found another in the bedroom later, and it was a very stressful night. I am wondering how I can control my phobia better. Or, if anyone has tips on how to keep critters at bay. I would also love if anyone could offer advice on how to come down from a panic attack. Ive had many in the past, and usually I can bring myself down after 510 minutes, but this time my husband had to talk me through it.
} & 0 & 1 & 1 \\
\cmidrule(lr){1-4}
\textit{Does anyone else feel like theyre antidepressants made an asshole. Dont get me wrong, \textcolor{red}{this is the best Ive felt in a long time}. Im actually capable of getting up and brushing my teeth and showering without needing reminded like a child. Ive been meal prepping my lunches for work and \textcolor{red}{I feel pretty great. But I know for a fact before I was diagnosed with the depression} I had a greater sense of empathy for people. I loved working in retail because I felt like I had a better understanding for why people were in bad moods. Now that I feel like Ive leveled out more, I just dont wanna tolerate peoples shit. Before, if someone came in and they were mean, I could justify that maybe theyve been having a bad day and Im just the five minute encounter they took it out on. Ive considered going off the antidepressants just because Im not sure if I like who Ive turned into. TLDR Had more empathy for people when I was more depressed. Getting better is turning me into an impatient bitch?} & 0 & 1 & 1\\
\cmidrule(lr){1-4}
\textit{'\textcolor{red}{I am hopeful about recovery}. Hey guys,I was **diagnosed with depression and anxiety in 2009** and are on medication for both of these issue. I am still taking low doses of meds but my breakthrough year was in 2015 where I started to meet different people. \textcolor{red}{For the past week or to be exact 8 days I have been feeling very well mentally},when I feel depression or anxiety creeping in I would call my friend and talk to them,or even call the suicide hotline for help to calm myself down.If I can continue to do well and not get over stress I am confident that if I can maintain all thesegood sleep,socializing,going to work,NoFapI will be able to live a better life in the future. The thing I feel hopeful is that if all these is maintained,I would be able to find a GF,because my mental health has been the main obstacles in dating.I am very thankful for life right now,but I just hope god can let me continue to work towards recovery and continue to feel better. What helps me is these 1.Socializing 2.Sleep 3.Nofap 4.psychotherapy 5.Work 6.Calling people and talk’
} & 0 & 1 & 1\\
\bottomrule
\end{tabular}%
}
\caption{Depression classification comparison GPT3.5 vs GPT4 --  Sample 1 Temporal Limitations in Depression Assessment by Language Models. In sample 2, despite a participant's self-reported mood improvement, the model incorrectly infers remission from depression. Sample 3 illustrates how the model inaccurately concludes no depression based on recent positive self-reports.
}
\label{tab:sB1}
\end{table*}

\begin{table*}[!ht]
\centering
\resizebox{\textwidth}{!}{%
\begin{tabular}{p{8.5in}ccc}
\toprule
\textbf{Example} & \textbf{True} & \textbf{GPT3.5} & \textbf{GPT4} \\
\midrule
\textit{Anxiety disorder exacerbating phobias. Background late  Feeling so much better from the last time I've posted here. 4 months ago, I felt so down. I felt as if someone put gravity on the highest level and I was being pulled into the ground. Constant \textcolor{red}{thoughts of suicide} followed me everywhere. Work, school, when I was alone at night. I knew how bad it was getting and how badly I wanted it. Instead I got help. The reason being was seeing everyone's reaction after being sent to the mental health hospital. There was so much good vibes around me it inspired me. Also everyone at the hospital especially inspired me. There are people out that who are good and who only want to help. When I got out, I started seeing a therapist and taking medication. I also started doing things for myself more. Not going out as much, binge watching good shows, reading, learning new recipes. Your own company is the best and it's been the best for the last four months. Also I quit drinking and I've never regretted that decision. It gets better friends, hold in there.} & 0 & 1 & 1 \\
\cmidrule(lr){1-4}
\textit{Almost 6 days feeling great. Except for yesterday but I had BPD abandonment issues but I wasn't exactly depressed. I've tracked my mood for the past couple of days and the last time I felt depressed and sad was Jan. 9th! I don't wanna claim victory just yet but I guess my mood stabilizer + antidepressant have worked after a couple of months. Today my mother hugged me and told me she was very happy to see me this well after so long, and that she loves me, always has and always will. I'm so proud of these days except for yesterday lol and I know that you guys can get back up and keep it up 3 I thought \textcolor{red}{I was going to end up killing myself and actually tried to do it.} I'm glad I'm still here, although I'm a little bit scared for my third year of University like omg it's getting scary.} & 0 & 1 & 1\\
\bottomrule
\end{tabular}%
}
\caption{Examples illustrating the limitations of GPT-3.5 and GPT-4 in zero-shot anxiety classification. Text highlighted in red indicates suicidal traits incorrectly classified by the models as anxiety disorder.}
\label{tab:sB2}
\end{table*}

\begin{table*}[!ht]
\centering
\resizebox{\textwidth}{!}{%
\begin{tabular}{p{8.5in}ccc}
\toprule
\textbf{Example} & \textbf{True} & \textbf{GPT4} & \textbf{XL-NET} \\
\midrule
\textit{Having constricted breathinginability to catch your breathfeeling like you cant fill your lungs? Read THIS. Hey everyone, So anyone familiar with panic attacksanxiety will probably know the feeling of not being able to catch your breath, having constricted lungs, or the feeling you cant fill your lungs. Anyway if youve tried breathing techniques, tried all the naturalistic ways of treating it and nothing is working then hear me out Today I talked to my doctor and she had no problem prescribing me a \textcolor{red}{salbutamol ventolin inhaler}. Its a really quick release from the constrictedlimited breathing you typically feel with anxiety and the feeling of inability to catch your breathfill you lungs. I have only used it once and \textcolor{red}{it works AMAZING.} So far one use has lasted me a few hours still going and all my anxiety from \textcolor{red}{breathing feels suddenly lifted}. Also my prescription which came with 200 uses with a max of 2 uses per day, so this is a somewhat longterm solution. I am also on a few other things for anxietydepression but at this point I almost feel I dont need them because I have finally gotten control of my breathing. Anyway hope this helps, dont be afraid to talk to your doctors, people!
} & 1 & 0 & 0 \\
\cmidrule(lr){1-4}
\textit{Is smoking weed while on Zoloft a bad idea. Ive been \textcolor{red}{feeling a lot better thanks to Zoloft}. Life is going good and I feel comfortable and somewhat happy at times. I still have a lot of issues I need to work on which is why I go to therapy every two weeks. I sometimes think I should go once a week but I dont know. Im also on mirtazapine and risperidone. Is there anything wrong with** smoking weed** every week? Im not using it to cope or anything. I just use it because it is \textcolor{red}{fun and relaxing}. I just like to smoke and paint on canvases and listen to music. I know how easy it is to be addicted to it. I dont smoke every single second and every single day. I mostly just smoke late at night and chill. My therapist said that it can increase anxiety and depression. All my friends think Im fine and its not a big deal I smoke. But ganja these days is very powerful. And I sometimes worry about what its doing to my brain. Ive been smoking the past 4 weeks and havent had any issues. A few bad trips here and there when I got too high, but still, it was all very enjoyable. My psychiatrist was telling me how its dangerous because Im buying it from the streets. And that it could be laced and I might end up in the emergency room. Ive been smoking for the past 2 years at age 24 and never had anything laced or anything like that. I know smoking weed isnt a great habit. And its not really the greatest thing to be doing, but I just like to chill and draw in my sketchbook and listen to some music. Is that such a bad thing?
} & 1 & 0 & 0\\
\bottomrule
\end{tabular}%
}
\caption{Examples highlighting the limitations of GPT-4 and XL-NET in depression classification. Text highlighted in red indicates instances where the model might incorrectly classify individuals as non-depressed due to the influence of medication on their improved mood.}
\label{tab:sC}
\end{table*}

\subsection{Error-Analysis -- GPT4 vs Best PLM}
The analysis shows thatfor depression 86\% of samples yield the same output across BERT-based models. However, mentioning certain disorders often leads to misclassification as depression (Refer Table \ref{tab:sC}). GPT-4 tends to classify longer samples more accurately, while XL-NET struggles with them, as evident from table \ref{tab:char_cnt}.

\begin{table}[t!]
\centering
\resizebox{\columnwidth}{!}{%
\begin{tabular}{@{} >{\centering\arraybackslash}m{3cm} >{\centering\arraybackslash}m{3cm} >{\centering\arraybackslash}m{3cm} >{\centering\arraybackslash}m{3cm} @{}}
\toprule
\multicolumn{2}{c}{\textbf{Models}}         & \multicolumn{2}{c}{\textbf{Character Count}} \\
\cmidrule(lr){1-2} \cmidrule(lr){3-4}
XL-NET & GPT-4 & ND     & D      \\
\midrule
T      & T     & $999.2$  & $1057.3$ \\
T      & F     & $979.6$  & $997.7$  \\
F      & T     & $1005.5$ & $1021.3$ \\
F      & F     & $987.2$  & $1040$   \\
\cmidrule(lr){1-2} \cmidrule(lr){3-4}
\multicolumn{2}{c}{\textbf{Ground Truth}}   & 992    & 1052   \\
\bottomrule
\end{tabular}%
}
\caption{Effect of Character Count on Model Classification: T denotes correctly classified samples, while F denotes incorrectly classified samples. ND indicates non-depressed posts, and D indicates depressed posts}
\label{tab:char_cnt}
\end{table}

\begin{table*}[!ht]
\begin{center}
\small 
\setlength{\tabcolsep}{2pt} 
\renewcommand{\arraystretch}{1.0} 
\resizebox{\textwidth}{!}{%
\begin{tabular}{p{6.5in}}
\toprule
\texttt{Reddit Post: <post> }\\\\
\texttt{Does the language of the given Reddit post indicate any cues relating to some kind of mental health disorder? Refer to DSM5 to justify your response.}\\\\
\texttt{{Strictly follow the following response format:}}\\\\
\texttt{Mental Health Disorder: (Yes/No)}\\\\
\texttt{Name of Mental Health Disorder:}\\\\
\texttt{DSM5 Rationale:}\\\\
\bottomrule
\end{tabular}%
}
\end{center}
\caption{Open-ended evaluation prompt to get Silver Labels from a Reddit post based on DSM5 criteria.}
\label{tab:silver_label_prompt}
\end{table*}

\begin{table*}[!ht]
\begin{center}
\small 
\setlength{\tabcolsep}{2pt} 
\renewcommand{\arraystretch}{1.0} 
\resizebox{\textwidth}{!}{%
\begin{tabular}{p{6.5in}}
\toprule
\texttt{Below are the 10 symptoms of depression and their definitions according to the Montgomery and Åsberg Depression Rating Scale. As a psychologist, read the social media post and evaluate the severity score of each symptom on a scale of 0 to 6. The social media post is delimited with triple backticks. Format your response as a JSON object with serial number of the 10 symptoms as keys. Do not return anything else apart from the json object.}\\\\
\texttt{\textbf{1. Apparent Sadness:} Representing despondency, gloom and despair, (more than just ordinary transient low spirits) reflected in speech, facial expression, and posture.}\\\\
\texttt{\textbf{2. Reported sadness:} Representing reports of depressed mood, regardless of whether it is reflected in appearance or not. Includes low spirits, despondency or the feeling of being beyond help and without hope.}\\\\
\texttt{\textbf{3. Inner tension:} Representing feelings of ill-defined discomfort, edginess, inner turmoil, mental tension mounting to either panic, dread or anguish.}\\\\
\texttt{\textbf{4. Reduced sleep:} Representing the experience of reduced duration or depth of sleep compared to the subject's own normal pattern when well.}\\\\
\texttt{\textbf{5. Reduced appetite:} Representing the feeling of a loss of appetite compared with when well.}\\\\
\texttt{\textbf{6. Concentration Difficulties:} Representing difficulties in collecting one's thoughts mounting to incapacitating lack of concentration. Rate according to intensity, frequency, and degree of incapacity produced.}\\\\
\texttt{\textbf{7. Lassitude:} Representing a difficulty getting started or slowness initiating and performing everyday activities.}\\\\
\texttt{\textbf{8. Inability to feel:} Representing the subjective experience of reduced interest in the surroundings, or activities that normally give pleasure. The ability to react with adequate emotion to circumstances or people is reduced.}\\\\
\texttt{\textbf{9. Pessimistic thoughts:} Representing thoughts of guilt, inferiority, self-reproach, sinfulness, remorse and ruin.}\\\\
\texttt{\textbf{10. Suicidal thoughts:} Representing the feeling that life is not worth living, that a natural death would be welcome, suicidal thoughts, and preparations for suicide.}\\\\
\texttt{Post:} \\```\\
\bottomrule
\end{tabular}%
}
\end{center}
\caption{Prompt template used for evaluating depression severity from social media posts based on MADRS scale.}
\label{tab:DEPRESSION_SEVERITY_PROMPT}
\end{table*}

\begin{table*}[!ht]
\begin{center}
\small 
\setlength{\tabcolsep}{2pt} 
\renewcommand{\arraystretch}{1.0} 
\resizebox{\textwidth}{!}{%
\begin{tabular}{p{6.5in}}
\toprule
\texttt{Below are the 9 symptoms of depression according to the Patient Health Questionnaire (PHQ-9). As a psychologist, read the social media post and determine which of the 9 symptoms observed in the post. The social media post is delimited with triple backticks. Format your response as a JSON object with serial number of the 9 symptoms as keys and "yes" or "no" as values. Do not return anything else apart from the json object.}\\\\
\texttt{{1. Little interest or pleasure in doing things:} }\\\\
\texttt{{2. Feeling down, depressed, or hopeless.} }\\\\
\texttt{{3. Trouble falling or staying asleep, or sleeping too much.} }\\\\
\texttt{{4. Feeling tired or having little energy.} }\\\\
\texttt{{5. Poor appetite or overeating.} }\\\\
\texttt{{6. Feeling bad about yourself – or that you are a failure or have let yourself or your family down.} }\\\\
\texttt{{7. Trouble concentrating on things, such as reading the newspaper or watching television.} }\\\\
\texttt{{8. Moving or speaking so slowly that other people could have noticed. Or the opposite – being so fidgety or restless that you have been moving around a lot more than usual.} }\\\\
\texttt{{9. Thoughts that you would be better off dead, or of hurting yourself in some way.} }\\\\
\texttt{Post:} \\```\\
\bottomrule
\end{tabular}%
}
\end{center}
\caption{Prompt template used for evaluating depression symptoms from social media posts based on PHQ-9.}
\label{tab:PHQ9_PROMPT}
\end{table*}

\begin{table*}[!ht]
\begin{center}
\small 
\setlength{\tabcolsep}{2pt} 
\renewcommand{\arraystretch}{1.0} 
\resizebox{\textwidth}{!}{%
\begin{tabular}{p{6.5in}}
\toprule
\texttt{Below are the 14 symptoms of anxiety and their definitions according to the Hamilton Anxiety Rating Scale. As a psychologist, read the social media post and evaluate the severity score of each symptom on a scale of 0 to 4. The social media post is delimited with triple backticks. Format your response as a JSON object with serial number of the 14 symptoms as keys. Do not return anything else apart from the json object.}\\\\
\texttt{\textbf{1. Anxious mood:} Worries, anticipation of the worst, fearful anticipation, irritability.}\\\\
\texttt{\textbf{2. Tension:} Feelings of tension, fatigability, startle response, moved to tears easily, trembling, feelings of restlessness, inability to relax.}\\\\
\texttt{\textbf{3. Fears:} Of dark, of strangers, of being left alone, of animals, of traffic, of crowds.}\\\\
\texttt{\textbf{4. Insomnia:} Difficulty in falling asleep, broken sleep, unsatisfying sleep and fatigue on waking, dreams, nightmares, night terrors.}\\\\
\texttt{\textbf{5. Intellectual:} Difficulty in concentration, poor memory.}\\\\
\texttt{\textbf{6. Depressed mood:} Loss of interest, lack of pleasure in hobbies, depression, early waking, diurnal swing.}\\\\
\texttt{\textbf{7. Somatic (muscular):} Pains and aches, twitching, stiffness, myoclonic jerks, grinding of teeth, unsteady voice, increased muscular tone.}\\\\
\texttt{\textbf{8. Somatic (sensory):} Tinnitus, blurring of vision, hot and cold flushes, feelings of weakness, pricking sensation.}\\\\
\texttt{\textbf{9. Cardiovascular symptoms:} Tachycardia, palpitations, pain in chest, throbbing of vessels, fainting feelings, missing beat.}\\\\
\texttt{\textbf{10. Respiratory symptoms:} Pressure or constriction in chest, choking feelings, sighing, dyspnea.}\\\\
\texttt{\textbf{11. Gastrointestinal symptoms:} Difficulty in swallowing, wind abdominal pain, burning sensations, abdominal fullness, nausea, vomiting, borborygmi, looseness of bowels, loss of weight, constipation.}\\\\
\texttt{\textbf{12. Genitourinary symptoms:} Frequency of micturition, urgency of micturition, amenorrhea, menorrhagia, development of frigidity, premature ejaculation, loss of libido, impotence.}\\\\
\texttt{\textbf{13. Autonomic symptoms:} Dry mouth, flushing, pallor, tendency to sweat, giddiness, tension headache, raising of hair.}\\\\
\texttt{\textbf{14. Behavior at interview:} Fidgeting, restlessness or pacing, tremor of hands, furrowed brow, strained face, sighing or rapid respiration, facial pallor, swallowing, etc.}\\\\
\texttt{Post:} \\```\\
\bottomrule
\end{tabular}%
}
\end{center}
\caption{Prompt template used for evaluating anxiety symptoms from social media posts based on the Hamilton Anxiety Rating Scale.}
\label{tab:HARS_PROMPT}
\end{table*}

\begin{table*}[!ht]
\begin{center}
\small 
\setlength{\tabcolsep}{2pt} 
\renewcommand{\arraystretch}{1.0} 
\resizebox{\textwidth}{!}{%
\begin{tabular}{p{6.5in}}
\toprule
\texttt{Below are 21 common symptoms of anxiety according to the Beck Anxiety Inventory (BAI). As a psychologist, read the social media post and evaluate the severity score of each symptom on a scale of 0 to 4. The social media post is delimited with triple backticks. Format your response as a JSON object with serial number of the 21 symptoms as keys.}\\\\
\texttt{\textbf{1. Numbness or tingling}}\\\\
\texttt{\textbf{2. Feeling hot}}\\\\
\texttt{\textbf{3. Wobbliness in legs}}\\\\
\texttt{\textbf{4. Unable to relax}}\\\\
\texttt{\textbf{5. Fear of the worst happening}}\\\\
\texttt{\textbf{6. Dizzy or lightheaded}}\\\\
\texttt{\textbf{7. Heart pounding/racing}}\\\\
\texttt{\textbf{8. Unsteady}}\\\\
\texttt{\textbf{9. Terrified or afraid}}\\\\
\texttt{\textbf{10. Nervous}}\\\\
\texttt{\textbf{11. Feeling of choking}}\\\\
\texttt{\textbf{12. Hands trembling}}\\\\
\texttt{\textbf{13. Shaky/unsteady}}\\\\
\texttt{\textbf{14. Fear of losing control}}\\\\
\texttt{\textbf{15. Difficulty in breathing}}\\\\
\texttt{\textbf{16. Fear of dying}}\\\\
\texttt{\textbf{17. Feeling scared}}\\\\
\texttt{\textbf{18. Indigestion}}\\\\
\texttt{\textbf{19. Faint/lightheaded}}\\\\
\texttt{\textbf{20. Face flushed}}\\\\
\texttt{\textbf{21. Hot/cold sweats}}\\\\
\texttt{Post:} \\```\\
\bottomrule
\end{tabular}%
}
\end{center}
\caption{Prompt template used for evaluating anxiety symptoms from social media posts based on BAI.}
\label{tab:BAI_PROMPT}
\end{table*}

\end{document}